\documentclass[11pt]{article}
\usepackage{times}

\usepackage[utf8]{inputenc}
\usepackage{blindtext}

\newtheorem{criterion}{Criterion}

\usepackage{mdframed}

\usepackage{multirow}
\usepackage{booktabs}
\usepackage{comment}
\usepackage{amssymb}
\usepackage[section]{placeins}

\usepackage{longtable}
\usepackage{makecell}

\usepackage{graphicx}
\usepackage[numbers,sort&compress]{natbib}

\usepackage[margin=1in]{geometry}
\usepackage[compact]{titlesec}
\titlespacing{\section}{0pt}{0pt}{0pt}
\usepackage{authblk}

\title{Out of One, Many:\\Using Language Models to Simulate Human Samples}

\author[1]{Lisa P. Argyle}
\author[1]{Ethan C. Busby}
\author[2]{Nancy Fulda}
\author[1]{Joshua Gubler}
\author[2]{Christopher Rytting}
\author[2]{David Wingate}

\affil[1]{Department of Political Science, Brigham Young University}
\affil[2]{Department of Computer Science, Brigham Young University}

\begin{document}

\maketitle

\begin{abstract}
We propose and explore the possibility that language models can be studied as effective proxies for specific human sub-populations in social science research. Practical and research applications of artificial intelligence tools have sometimes been limited by problematic biases (such as racism or sexism), which are often treated as uniform properties of the models. We show that the “algorithmic bias” within one such tool-- the GPT-3 language model-- is instead both fine-grained and demographically correlated, meaning that proper conditioning will cause it to accurately emulate response distributions from a wide variety of human subgroups.  We term this property \textit{algorithmic fidelity} and explore its extent in GPT-3. We create ``silicon samples'' by conditioning the model on thousands of socio-demographic backstories from real human participants in multiple large surveys conducted in the United States. We then compare the silicon and human samples to demonstrate that the information contained in GPT-3 goes far beyond surface similarity. It is nuanced, multifaceted, and reflects the complex interplay between ideas, attitudes, and socio-cultural context that characterize human attitudes.  We suggest that language models with sufficient algorithmic fidelity thus constitute a novel and powerful tool to advance understanding of humans and society across a variety of disciplines.

\end{abstract}

\tableofcontents

\parindent 0.0in
\parskip 0.15in

\section{Introduction}

Recent years have witnessed a marked increase in the use of machine learning tools to advance social science research \citep{rodriguez2021word, benoit2019measuring, barbera2021automated, rheault2020word, grimmer2021machine, greene2019machine}. However, little attention has yet been paid to the possible applications of large-scale generative language models like GPT-2 \citep{radford2019language}, T5 \citep{2020t5}, or GPT-3 \citep{brown2020language} to advancing scientific understanding of human social and political behavior. These models are complex conditional distributions over natural language that are used to generate synthetic text. When trained at scale, they exhibit a remarkable ability to capture patterns of grammar, cultural knowledge, and conversational rhythms present in natural language \citep{Meena, openaiGPT2, dai2019transformerxl}, and have become so convincing that the texts they generate are largely indistinguishable from those generated by humans \citep{brown2020language}. We propose that these models can be used as surrogates for human respondents in a variety of social science tasks.

Our argument begins with a different take on a commonly-recognized problem with artificial intelligence tools: their penchant for replicating the racial, gender, economic, and other biases of their creators.
Most discussions of this ``algorithmic bias'' treat it as a singular, macro-level feature of the model, and seek ways to mitigate negative effects \citep{panchArtificialIntelligenceAlgorithmic2019, maysonBiasBiasOut2018, barocasBigDataDisparate2016}. We suggest it is better understood as a complex reflection of the many various patterns of association between ideas, attitudes, and contexts present among humans. Our studies show that the same language model, when properly conditioned, is able to produce outputs biased both toward \textit{and} against specific groups and perspectives in ways that strongly correspond with human response patterns along fine-grained demographic axes. In other words, these language models do not contain just one bias, but \textit{many}. This means that by conditioning a model on simulated ``individuals'' with targeted identity and personality profiles, it is possible to select from among a diverse and frequently disjoint set of response distributions within the model, each closely aligned with a real human sub-population. We call the degree to which a model can accurately reflect these distributions its degree of \emph{algorithmic fidelity}.

High algorithmic fidelity in language models is crucial for their use in social science as it enables researchers to extract information from a single language model that provides insight into the different patterns of attitudes and ideas present across many groups (women, men, White people, people of color, millennials, baby boomers, etc.) and also the \emph{combination and intersection} of these groups (Black immigrants, female Republicans, White males, etc.). As yet, however, the extent of algorithmic fidelity in large-scale language models is unexplored. In three studies, we provide evidence that the GPT-3 language model \citep{brown2020language} satisfies what we argue are the four essential criteria of algorithmic fidelity. We obtain this evidence by conditioning GPT-3 on thousands of socio-demographic backstories from real human participants in multiple large surveys in the United States: the 2012, 2016, and 2020 waves of the American National Election Studies (ANES)\citep{anes}, and Rothschild et al.'s ``Pigeonholing Partisans'' data \citep{busby2019}. We condition the model to generate one ``silicon subject'' for each human study participant, and then ask these simulated subjects to complete the same tasks that were required of human participants. To assess algorithmic fidelity, we explore the degree to which the complex patterns of relationships between ideas, attitudes, and contexts within our silicon subjects accurately mirror those within the human populations. The results from our tests provide the first extensive, systematic exploration of the degree of algorithmic fidelity in a large-scale language model, laying the groundwork for the use of these models broadly in social science.

These studies also provide initial examples of just a few of the myriad potential ways language models can be used in social science research once algorithmic fidelity in a given domain is established. In Study 1, we ask our GPT-3 surrogates to list words describing outgroup partisans and show how closely these words mirror those listed by their human counterparts. In Studies 2 and 3, we explore the relationships between various demographics, attitudes, and reported behaviors; our results show the same patterns of relationships among GPT-3 surrogates and their human counterparts. For all three of these studies, we explain how a researcher might use only the information from GPT-3 to more effectively study human populations. These results suggest that in the realm of U.S. politics, researchers can confidently use a GPT-3 ``silicon sample'' to explore hypotheses prior to costly deployment with human subjects. GPT-3 can thus be used both in theory generation and testing. 


This paper makes five important contributions: (1) it conceptualizes algorithmic fidelity and identifies four criteria to assess it; (2) it introduces \emph{silicon sampling}, a methodology by which a language model can generate a virtual population of respondents, correcting skewed marginal statistics in the training data; (3) it introduces a novel approach to conditioning on first-person demographic backstories to simulate targeted human survey responses; (4) it presents compelling evidence for the existence of algorithmic fidelity in the GPT-3 language model in the domain of U.S. politics and public opinion; and (5) it provides examples of how the model can be used for social science research in this domain. 


\section{The GPT-3 Language Model}


The GPT-3 language model holds particular promise as a social science tool. Released by OpenAI in 2020, GPT-3 has 175 billion parameters and was trained on more than 45 terabytes of text, making it one of the largest generative language models ever created. Texts generated by GPT-3 are strikingly difficult to distinguish from authentic human compositions.

Formally, language models like GPT-3 are a conditional probability distribution $p(x_n|x_1,\cdots,x_{n-1})$ over tokens,
where each $x_i$ comes from a fixed vocabulary.
By iteratively sampling from this distribution, a language model can generate arbitrarily long sequences of text.
However, before it can generate text, a language model like GPT-3 requires ``conditioning,'' meaning that it must be provided with initial input tokens comprising $\{x_1, ..., x_{n-1}\}$. We refer to this conditioning text as the model's \emph{context}.

Conditioning a language model on different contexts reduces the probability of some outputs and increases the probability of others. For example, given the context $\{x_1,x_2,x_3\}=$``Can you come'', a language model might assign high probability to $x_4$=``home'', and low probability to $x_4$=``bananas'', but changing a single word in the context to $\{x_1,x_2,x_3\}=$ ``Can you eat'' reverses that. At each generative step, the model estimates a probability distribution corresponding to the likelihood that any given token in the vocabulary would have been the next observed $x_i$ if the model were reading a pre-written text. Using a distribution function, it selects one of the most probable candidates, the new $x_i$ is appended to the conditioning context, and the entire process repeats. This continues until a pre-specified number of tokens has been generated, or until an external factor causes the process to stop. Because GPT-3 selects output tokens probabilistically, it can generate many possible continuations for a given context. 

\section{Algorithmic Fidelity}

We define \textit{algorithmic fidelity} as the degree to which the complex patterns of relationships between ideas, attitudes, and socio-cultural contexts within a model accurately mirror those within a range of human sub-populations. The core assumption of algorithmic fidelity is that texts generated by the model are selected not from a single overarching probability distribution, but from a combination of \textit{many} distributions, and that structured curation of the conditioning context can induce the model to produce outputs that correlate with the attitudes, opinions, experiences of distinct human sub-populations.

This does not imply that the model can simulate a specific individual or that every generated response will be coherent. Many of the known shortcomings and inaccuracies of large language models still apply \citep{bender2021dangers, marcus2020next}. However, by selecting a conditioning context that evokes the shared socio-cultural experience of a specific demographic group, we find that it is possible to produce response distributions that strongly correlate with the distribution of human responses to survey questions from that demographic.

Our conception of algorithmic fidelity goes beyond prior observations that language models reflect human-like biases present in the text corpora used to create them \citep{garg2018word, caliskan2017semantics, panchArtificialIntelligenceAlgorithmic2019, maysonBiasBiasOut2018, barocasBigDataDisparate2016}. Instead, it suggests that the high-level, human-like output of language models stems from human-like underlying concept associations. This means that given basic human demographic background information, the model exhibits underlying patterns between concepts, ideas, and attitudes that mirror those recorded from humans with matching backgrounds. To use terms common to social science research, algorithmic fidelity helps to establish the generalizability of language models, or the degree to which we can apply what we learn from language models to the world beyond those models. 


How much algorithmic fidelity in a language model is \textit{enough} for social science use? We suggest at a minimum a language model must provide repeated, consistent evidence of meeting the following four criteria:


\begin{criterion}
\textbf{(Social Science Turing Test)} Generated responses are indistinguishable from parallel human texts.
\end{criterion}

\begin{criterion}
\textbf{(Backward Continuity)} Generated responses are consistent with the attitudes and socio-demographic information of its input/``conditioning context,'' such that  humans viewing the responses can infer key elements of that input.
\end{criterion}

\begin{criterion}
\textbf{(Forward Continuity)} Generated responses proceed naturally from the conditioning context provided, reliably reflecting the form, tone, and content of the context.
\end{criterion}

\begin{criterion}
\textbf{(Pattern Correspondence)}
Generated responses reflect underlying patterns of relationships between ideas, demographics, and behavior that would be observed in comparable human-produced data.
\end{criterion}


These criteria represent four qualitatively different dimensions on which a model must have fidelity to human responses if researchers are to have confidence in the ability of the model to generate reliable surrogate human responses. A lack of fidelity in any one of these four areas decreases confidence in its usability; a lack of fidelity in more than one decreases confidence further. We do not propose specific metrics or numerical thresholds to quantify meeting or missing these criteria, as the appropriate statistics will depend on varying data structures and disciplinary standards. Instead,we suggest the best metric is repeated support for each criteria across multiple data sources, different measures, and across many groups.

In the following studies, we take this approach in examining support for these criteria within GPT-3 in the domain of U.S. politics and public opinion. Decades of research in political science provide a robust literature identifying expected relationship patterns between political ideas, concepts, and attitudes \citep{Berelson1954, Campbell1960, HutchingsValentino2004, BurnsGallagher2010, DruckmanLupia2016, Cramer2020}. We leverage these as a basis for comparison.


\section{Silicon Sampling: Correcting Skewed Marginals}
\label{sec:siliconsampling}

Applying language models to social science research raises an obvious question: how can we compensate for the fact that the demographics of internet users (on which the model was trained) are neither  representative of most populations of interest nor demographically balanced, and that language models are trained on internet snapshots acquired at a fixed point in time?

We propose a general methodology, which we term \emph{silicon sampling}, that corrects skewed marginal statistics of a language model. To see what needs correcting, imagine trying to use GPT-3 to assess marginal probabilities of voting patterns $P(V)$.
GPT-3 models both voting patterns $V$ and demographics $B_{\mathrm{GPT3}}$ jointly as $P(V,B_{\mathrm{GPT3}}){=}P(V|B_{\mathrm{GPT3}})P(B_{\mathrm{GPT3}})$.

However, the distribution of backstories $P(B_{\mathrm{GPT3}})$ does not match the distribution $P(B_{\mathrm{True}})$ in the populations of interest to most social scientists (say, among all voting-eligible citizens); without correction, conclusions about marginal voting patterns $P(V){=}\int_{B} P(V,B_{\mathrm{GPT3}})$ will be skewed by this difference.
To overcome this, we leverage the conditional nature of language models and sample backstories from a known, nationally representative sample (for example, the ANES) and then estimate $P(V)$ based on those ANES-sampled backstories.
This allows us to compute $P(V|B_{\mathrm{ANES}})P(B_{\mathrm{ANES}})$. As long as GPT-3 models the \emph{conditional} distribution $P(V|B)$ well, we can explore patterns in \textit{any} designated population.

The conditional nature of GPT-3's text completions creates a situation analogous to Simpson's Paradox \citep{Simpson1951}, in which the trends evident in a combined group do not reflect the trends of its composite distributions. Specifically, our silicon sampling method allows us to examine the distinct opinions, biases, and voting patterns of identified sub-populations, which can differ drastically from the patterns exhibited by generic (i.e. not demographically conditioned) GPT-3 text completions.
Of course, the ability to sample from GPT-3's component text distributions does not, in and of itself, guarantee that these distributions faithfully reflect the behavior of specific human sub-populations. For that, one must first examine the model's algorithmic fidelity with respect to both the domain of study and the demographic groups of interest. 

\section{Study 1: Free-form Partisan Text}

Our first examination of algorithmic fidelity in GPT-3 involves a silicon replication of Rothschild et al.'s ``Pigeonholing Partisans" data \citep{busby2019}. This survey asked respondents to list four words to describe both Republicans and Democrats. Rothschild et al.~find that people talk about partisans in different ways, focusing on traits, political issues, social groups, or a combination of all three. Further, people often talk about their own political party in more positive ways than the other party, in line with other research \citep{Iyengar2012,Mason2018}. In this first test, we ask whether GPT-3 can produce texts about partisans that are, in the aggregate, indistinguishable from words generated by humans.

\begin{figure*}
\includegraphics[width=1\linewidth]{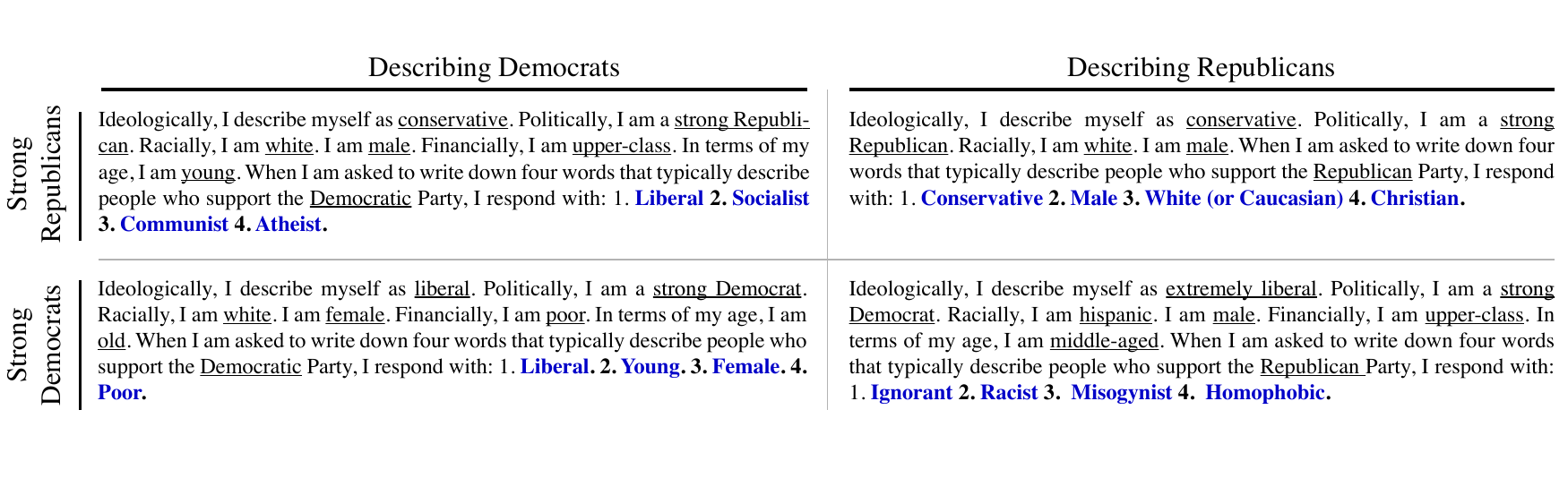}
\caption{Example contexts and completions from four silicon ``individuals'' analyzed in Study 1. Plaintext indicates the conditioning context; underlined words show demographics we dynamically inserted into the template; blue words are the four harvested words.}
\label{fig:pp_context}
\end{figure*}

To accomplish this, we leverage our silicon sampling technique, generating a synthetic dataset by constructing a first-person backstory for each human subject in the Pigeonholing Partisans survey, as shown in Figure \ref{fig:pp_context}. 
Using these texts, we ask GPT-3 to sample new words. Because of the setup of our conditioning text, GPT-3 almost always responds with neatly delineated sets of four words, although -- much like humans -- it occasionally responds with long phrases, mini-essays or nothing at all.  Post-processing with regular expressions extracted the final set of four words from each sample.

\begin{figure}[t!]
\centering
\includegraphics[width=.9\linewidth]{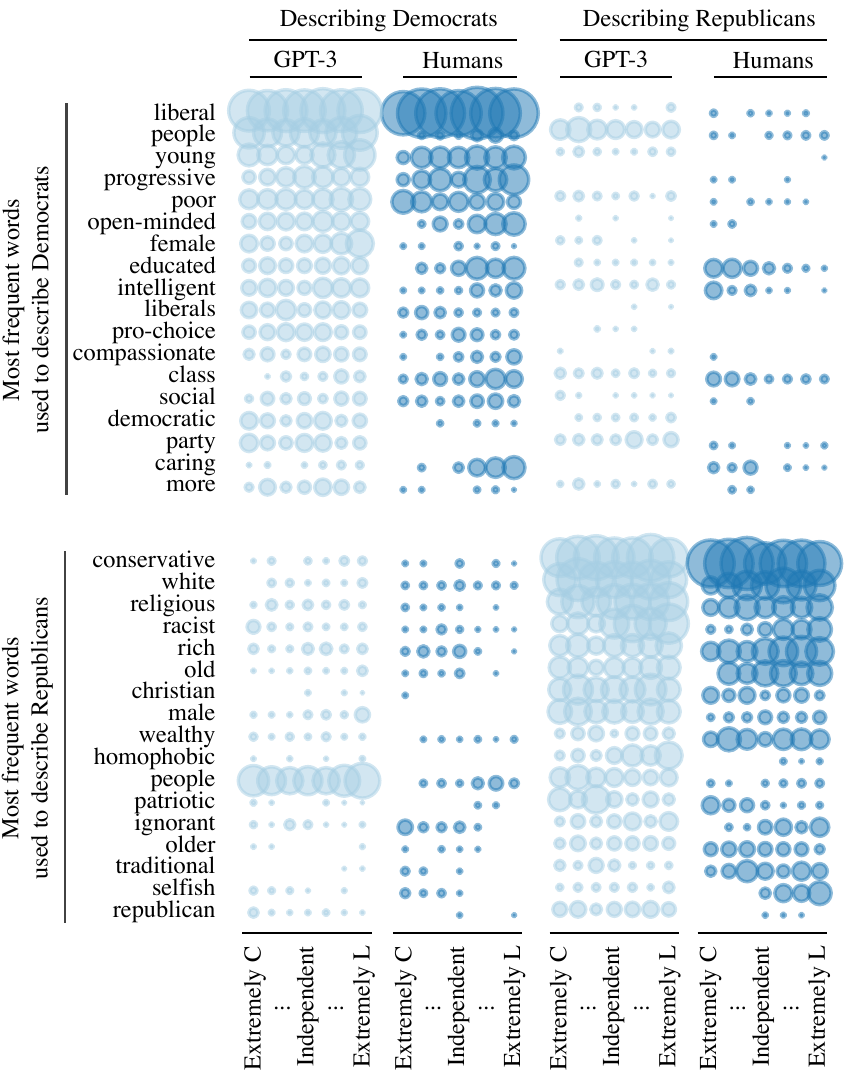}
\caption{The original Pigeonholing Partisans dataset and the corresponding GPT-3 generated words. Bubble size represents relative frequency of word occurrence; columns represent the ideology of list writers.  GPT-3 uses a similar set of words to humans.}
\label{fig:pp}
\end{figure}

For GPT-3 to generate 4-word lists that mirror human texts demands significant algorithmic fidelity, for it requires listing words that in tone and content mirror those listed by humans with a particular background. Figure \ref{fig:pp} compares the most frequent words used to describe Democrats and Republicans in our data, by data source (GPT-3 or human) and source ideology. Bubble size represents relative frequency of word occurrence; columns represent the ideology of the list writers. Qualitatively, both the human and GPT-3 lists look initially as political scientists might expect. For example, both GPT-3 and humans use a common set of words to describe Democrats, and rarely use those words to describe Republicans.

To formally analyze this data, we hired 2873 individuals through the survey platform \textit{Lucid} \citep{coppock2019} to evaluate the 7675 texts produced by human and GPT-3 survey respondents, without any indication of which was which. Each individual evaluated 8 randomly assigned lists, with each text evaluated by three different individuals.

We presented these evaluators with the 4-word lists after the following preface: ``Consider the following description of [Republicans/Democrats]:''. We then asked them to respond to six prompts. First, we asked them to guess the partisanship of the list writer (Republican, Democrat, or Independent). We then asked them to rate the list on 5 dimensions: (1) positive or negative tone, (2) overall extremity, and whether the text mentioned (3) traits, (4) policy issues, or (5) social groups.
Participants then sequentially viewed 8 additional randomly selected lists, were told that some of these lists were generated by a computer model, and were asked to guess whether each list was generated by a human or a computer. Extensive details on the lists, their writers, study participants, and the instructions can be found in the appendix.

Using this design, we explore two social science variations of a Turing Test: (1) whether our human evaluators recognize the difference between human and GPT-3-generated lists, and (2) whether the humans perceive the content of lists from both sources as similar. 
These tests speak to Criterion 1 (Turing test) and Criterion 2 (Backward Continuity).

We find evidence in favor of both criteria: participants guessed 61.7\% of human-generated lists were human-generated, while guessing the same of 61.2\% of GPT-3 lists (two-tailed difference p=0.44). Although asking participants to judge if a list is human- or computer-generated leads them to guess that some lists do not come from humans (nearly 40 percent of both kinds of lists fell in this category), this tendency does not vary by the source of the list.

This is particularly interesting given the results of our second exploration: whether participants noted any differences in list characteristics between human and GPT-3-generated lists. To identify these differences, we estimate regression models using ordinary least squares, regressing each of the 5 characteristics by which lists were evaluated (positivity, extremity, and mentions of traits, issues, and groups) on a dichotomous source variable (0 = human, 1 = GPT-3) and a series of control variables recording the gender, ethnicity, income, age, and partisan identity of the original list-writers in the Rothschild et al.~data. All models include fixed effects for evaluators (as each evaluated 8 lists), and clustered standard errors by evaluator and list (as each list was evaluated three times).

\begin{figure}[t!]
\centering
\includegraphics[width=.72\linewidth]{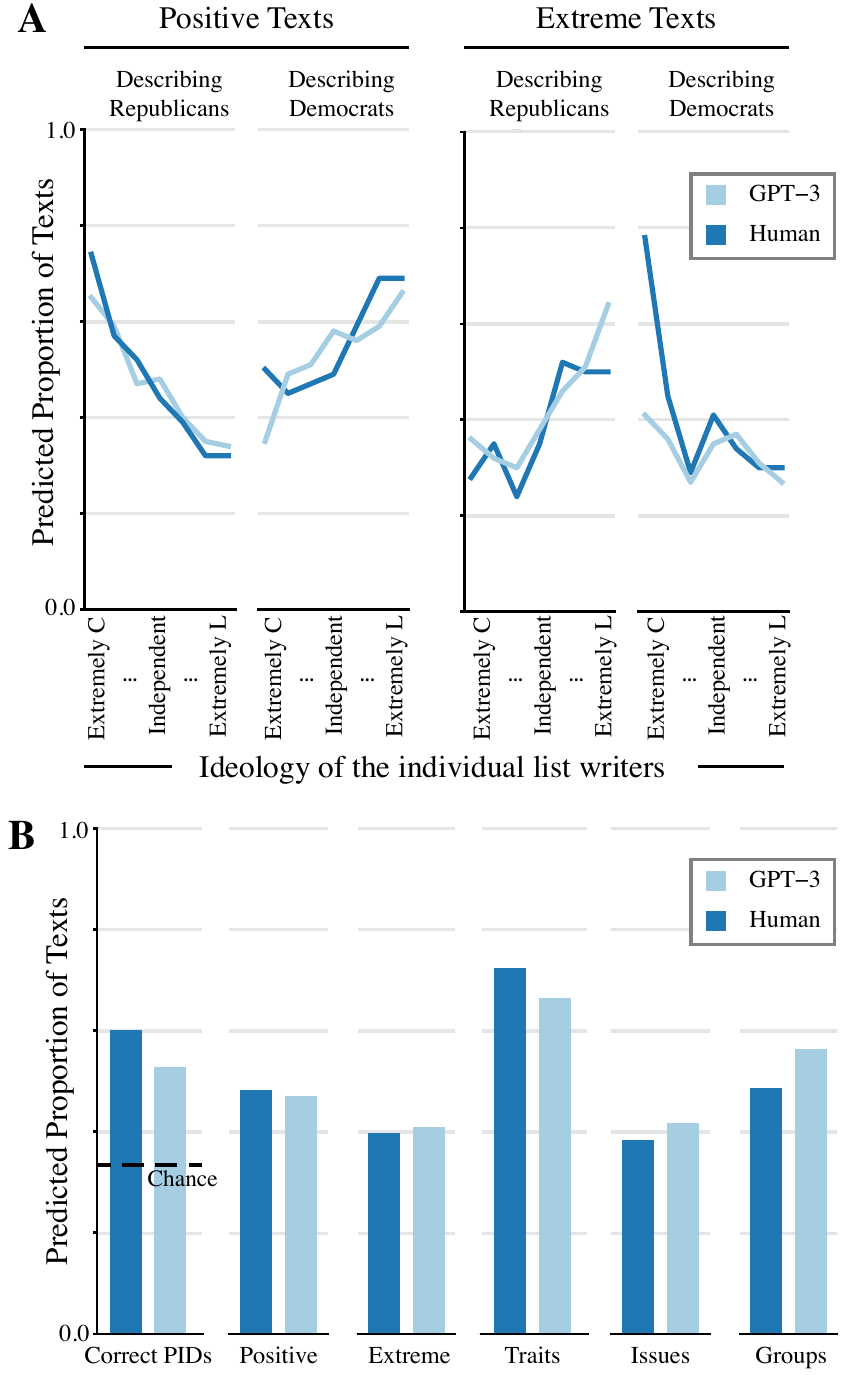}
\caption{Analysis of GPT-3 and human responses from the Lucid survey.}
\label{fig:lcomb}
\end{figure}

Figure \ref{fig:lcomb}(B) plots the predicted percent of all lists (human and GPT-3) evaluated as having each characteristic. The results show a remarkable degree of consistency in the evaluations of both human and GPT-3 generated lists in both content and tone. For example, human list-writers included more personality traits (e.g. ``bigoted,"``moral") than other components (72.3\% of lists). So did GPT-3 (66.5\% of lists). Less than half of both human and GPT-3 generated lists were evaluated as extreme (39.8\% and 41.0\%, respectively). This pattern of similarity holds across all 5 characteristics, with all but one characteristic hovering around 50\%. The lone exception, with a substantially higher frequency in both human and GPT-3 data, is ``traits." This matches patterns in the original analyses of the human texts \citep{busby2019}. That GPT-3 mirrors this exception, and patterns in all other characteristics, is strong evidence of the depth of algorithmic fidelity it contains. Tables of results and further model details can be found in the appendix.

Moreover, as Figure \ref{fig:lcomb}(A) indicates, when we drill down to greater levels of detail to explore underlying patterns behind these results, we find that GPT-3 reflects human-similar patterns at this level as well (Criterion 4, Pattern Correspondence). The similarity in the use of positive and extreme words by both humans and GPT-3, broken out by the ideological subgroup of the list writers, is striking.

We have shown that (1) human evaluators of the word lists could not correctly distinguish between human vs. GPT-3-generated lists, and (2) that they evaluated the content/characteristics of these lists as quite similar. We now assess the degree to which our participants were able to use these lists to correctly guess the true partisanship of the list writers. 
To explore this question, we estimate a model similar to those just presented, regressing a  dichotomous variable identifying if participants correctly guessed the partisanship of list writers (1 = Yes; 0 = No) on the source of the list (GPT-3 vs. human) and the same controls. The left-most bars of Figure \ref{fig:lcomb}(B) present the predicted percent correct, by source type.

Participants presented with lists of words from both sources guess the correct partisanship of their writer significantly better than chance (33\%, given respondents could guess Republican, Democrat, or Independent), providing strong additional evidence in favor of algorithmic fidelity in GPT-3. Participants who saw human-generated lists guessed successfully roughly 7.3\% more often (60.1\% vs. 52.8\%) than those who saw GPT-3 lists, a statistically significant difference (two-tailed $p<.001$). However, texts from both humans and GPT-3 both clearly contain the sentiment cues necessary to guess the partisanship of the creator of the texts at significant levels.

Results from study 1 suggest a remarkably high degree of algorithmic fidelity within GPT-3. We find repeated, consistent support for Criterion 1 (Turing Test) and Criterion 2 (Backward Continuity) from these data, with some initial evidence for Criterion 4 (Pattern Correspondence). In all of these cases, we observe support for these criteria across different measures and for different subsets of the American population. 

\section{Study 2: Vote Prediction}
Our next two studies draw on the ANES, a premier source of data in understanding American public opinion. 
In Study 2, we use the 2012, 2016, and 2020 waves of the survey.

We first consider the degree to which GPT-3 silicon samples, constructed to match the demographics of the 2012, 2016, and 2020 ANES participants, report similar distributions of vote choice as their matching human samples. This requires GPT-3 to generate text from a limited universe of options (vote for Trump or Clinton in 2016, for example), but it must do so differently based on the human backgrounds we provide it. It therefore assesses Criterion 3, requiring forward compatibility of GPT-3 in producing a candidate's name in response to the conditioning text, and Criterion 4, the expectation that the conditioning texts will produce predictable changes in the probability distribution of vote choice. To demonstrate algorithmic fidelity in this study, GPT-3 must contain human-like patterns of associations, where individuals from different ethnic groups, living in different locations, of different social backgrounds vote differently in ways identified but still debated by political scientists \citep{boxsteffensmeier2004, Tate1994, Cramer2016, Jardina2019}. Study 2 also explores the temporal limits of GPT-3: the training corpus for GPT-3 ended in 2019 \citep{brown2020language}, so data from 2020 allows us to explore how the algorithmic fidelity of the language model changes when probed outside the time of the original training corpus.

Drawing from research on voting in the United States, we used the following ANES variables to condition GPT-3: (1) racial/ethnic self-identification, (2) gender, (3) age, (4) conservative-liberal ideological self-placement, (5) party identification, (6) political interest, (7) church attendance, (8) if the respondent reported discussing politics with family and friends, (9) feelings of patriotism associated with the American flag (not available in 2020 at the time of analysis), and (10) state of residence (not available in 2020 at the time of the analysis). We then recorded the probabilities from GPT-3 that the sentence ``In [year], I voted for...'' would be completed with the Republican or Democratic candidate given the each conditioning backstory (see Online Appendix for further detail). Using these variables as conditioning text in GPT-3 allows us to compare how well a GPT-3 silicon sample replicates the relationships between each variable and vote choice in our human sample. In all the comparisons that follow, we code vote choice as 1 when the respondent/GPT-3 indicates a vote for the Republican candidate and a 0 for a vote for the Democratic candidate in that election. To make the predictions from GPT-3 match the observed human data, we dichotomized the probability predictions at 0.50, where higher values were scored as a vote for the Republican candidate.

\begin{table}[t!]
\small
\centering
\begin{tabular}{lc c c c c c}
\hline
\thead{Variable} & \thead{2012} & \thead{2012} & \thead{2016} &\thead{2016}& \thead{2020} &\thead{2020} \\
\thead{} & \thead{Tetra.} & \thead{Prop. Agree} & \thead{Tetra.} &\thead{Prop. Agree}& \thead{Tetra.} &\thead{Prop. Agree} \\
\hline
Whole sample & 0.90 & 0.85 & 0.92 & 0.87 & 0.94 & 0.89 \\
Men & 0.90 & 0.85 & 0.93 & 0.88 & 0.95 & 0.88 \\
Women & 0.91 & 0.86 & 0.92 & 0.86 & 0.94 & 0.90 \\
Strong partisans & 0.99 & 0.97 & 1.00 & 0.97 & 1.00 & 0.97 \\
Weak partisans & 0.73 & 0.74 & 0.71 & 0.74 & 0.84 & 0.82 \\
Leaners & 0.90 & 0.85 & 0.93 & 0.87 & 0.95 & 0.89 \\
Independents & 0.31 & 0.59 & 0.41 & 0.62 & 0.02 & 0.53 \\
Conservatives & 0.84 & 0.84 & 0.88 & 0.86 & 0.91 & 0.89 \\
Moderates & 0.65 & 0.77 & 0.76 & 0.78 & 0.71 &0.77 \\
Liberals & 0.81 & 0.95 & 0.73 & 0.95 & 0.86 & 0.97 \\
Whites & 0.87 & 0.82 & 0.91 & 0.85 &0.94 & 0.89 \\
Blacks & 0.71 & 0.97 & 0.87 & 0.96 & 0.81 & 0.94 \\
Hispanics & 0.86 & 0.86 & 0.93 & 0.90 & 0.88 & 0.83 \\
Attends church & 0.91 & 0.86 & 0.93 & 0.88 & 0.94 & 0.88 \\
Doesn't attend church & 0.88 & 0.85 & 0.90 & 0.85 &0.93 & 0.90 \\
High interest in politics & 0.95 & 0.90 & 0.97 & 0.93 & 0.97 &0.92 \\
Low interest in politics & 0.71 & 0.74 & 0.75 & 0.75 & 0.83 & 0.81\\
Discusses politics & 0.92 & 0.87 & 0.94 & 0.88 & 0.95 & 0.90 \\
Doesn't discuss politics & 0.83 & 0.82 & 0.81 & 0.79 & 0.80 & 0.79 \\
18 to 30 years old & 0.90 & 0.87 & 0.90 & 0.86 & 0.90 & 0.87 \\
31 to 45 years old & 0.90 & 0.85 & 0.92 & 0.87 & 0.94 & 0.90 \\
46 to 60 years old & 0.90 & 0.86 & 0.92 & 0.86 & 0.92 & 0.87\\
Over 60 & 0.90 & 0.85 & 0.93 & 0.87 & 0.96 & 0.91\\
\hline
\end{tabular}
\caption{Measures of correlation between GPT-3 and ANES probability of voting for the Republican presidential candidate. Tetra refers to tetrachoric correlation. Prop. Agree refers to proportion agreement. GPT-3 vote is a binary version of GPT-3's predicted probability of voting for the Republican candidate, dividing predictions at 0.50.}
\label{tab:anescorr}
\end{table}

We observe a high degree of correspondence between reported two-party presidential vote choice proportions from GPT-3 and ANES respondents. Averaged across the whole sample, GPT-3 reported a 0.391 probability of voting for Mitt Romney in 2012; the same percentage from the ANES was 0.404. In the 2016 data, GPT-3 estimated a 0.432 probability of voting for Trump, and the probability from the 2016 ANES was 0.477. In 2020, the GPT-3-generated probability of voting for Trump was 0.472, while the percentage from the ANES respondents was 0.412. In all three cases, we see evidence of a mild amount of overall bias in GPT-3: GPT-3 was a little predisposed against Romney in 2012, against Trump in 2016, and against Biden in 2020. However, the substantive difference between the ANES and GPT-3 estimates is relatively small and, in keeping with our larger arguments about algorithmic fidelity and corrections for skewed marginals, does not preclude strong and consistent correlations between GPT-3's simulated responses and the reactions of subgroups in the American population.

To explore these correlations in detail, we turn to the statistics reported in Table \ref{tab:anescorr}. This table reports two forms of correlations between the self-report of voting from the ANES and a binary version of the vote report from GPT-3 (other metrics support these two and can be found in the appendix). We dichotomize the GPT-3 vote probability to match our human measure, a binary report of voting from the ANES. Across all three years of survey data, we see remarkable correspondence between GPT-3 and human respondents. The 2012 tetrachoric correlation across all respondents 0.90, the 2016 estimate was 0.92, and the 2020 value was 0.94. We find this consistently high correlation remarkable given the differences in context across years.

This same high degree of pattern correspondence occurs for various subgroups in the American population. \textit{More than half} of the tetrachoric correlations between the reported vote by GPT-3 and the ANES are 0.90 or higher, and this is true for all three years. The proportion agreement column of Table \ref{tab:anescorr} also indicates high levels of raw agreement between the two reports of vote choice in 2012, 2016, and 2020. Impressively, there is only one exception to this overall pattern: the estimates of vote choice do not match well for pure independents, especially in 2020. However, this is the only deviation from the overall trend in Table \ref{tab:anescorr}, where all other measures of correspondence exceed 0.65 (and are generally closer to 0.8 or 0.9). Further, existing political science research suggests that this group of individuals should be especially hard to predict as they are the most conflicted about the two-party choices, the least likely to vote, the least politically knowledgeable, and the least interested in politics \citep{Keith1992, magleby2011, klar2016}. Overall, then, the results in Table \ref{tab:anescorr} provide strong, additional evidence for algorithmic fidelity, with repeated, consistent support for Criteria 3 (Forward Continuity) and 4 (Pattern Correspondence).
Appendix 3 contains additional results, including an ablation study investigating the effect of removing backstory elements, and a model comparison showing how alternative language models perform on this task.

The ability of GPT-3 to capture the voting preferences of different groups of Americans is not restricted to one moment in time. Moreover, results from the 2020 ANES data indicate the possibility that GPT-3 can be used, with the right conditioning, to understand people and groups outside its original training corpus. 

\section{Study 3: Closed-ended Questions and Complex Correlations in Human Data}

Study 3 examines GPT-3's ability to replicate complex patterns of association between a wide variety of conceptual nodes. Given the complexity of this task, we conduct it just for the 2016 data from the ANES. Building on the voting predictions in Study 2, we expand the set of information outputs we ask GPT-3 to produce, and use the resulting data to evaluate a more complex structure of associations. This is our most rigorous evaluation of Criterion 4 (Pattern Correspondence).

This study represents both a technical and substantive research challenge. Absent the naturally self-limiting set of likely responses when asking about vote choice in a particular election (i.e. ``Donald Trump'' vs.~``Hillary Clinton''), we develop a method to condition GPT-3 to provide specific responses from a list of options. Additionally, mirroring the widespread and varied use of survey data in social science applications, we push beyond high-level conditional probabilities and explore whether GPT-3 demonstrates algorithmic fidelity in inter-relationships among a variety of underlying attitudes, demographics, and reported behaviors.

For this task, we produce an interview-style conditioning template (see the appendix for an example). The purpose of this approach is two-fold. First, leveraging the zero-shot learning property of language models \citep{brown2020language}, the format induces GPT-3 to respond to survey questions using short strings of tokens drawn from options provided by the ``Interviewer.'' Second, the questions incorporated in the conditioning text provide necessary demographic and attitudinal background information to generate each distinct silicon subject. We generate the conditioning text using responses that humans gave on the 2016 ANES to eleven survey questions. We then use GPT-3 to predict the response to the twelfth.

\begin{figure*}[t!]
\centering
\includegraphics[width=1\linewidth]{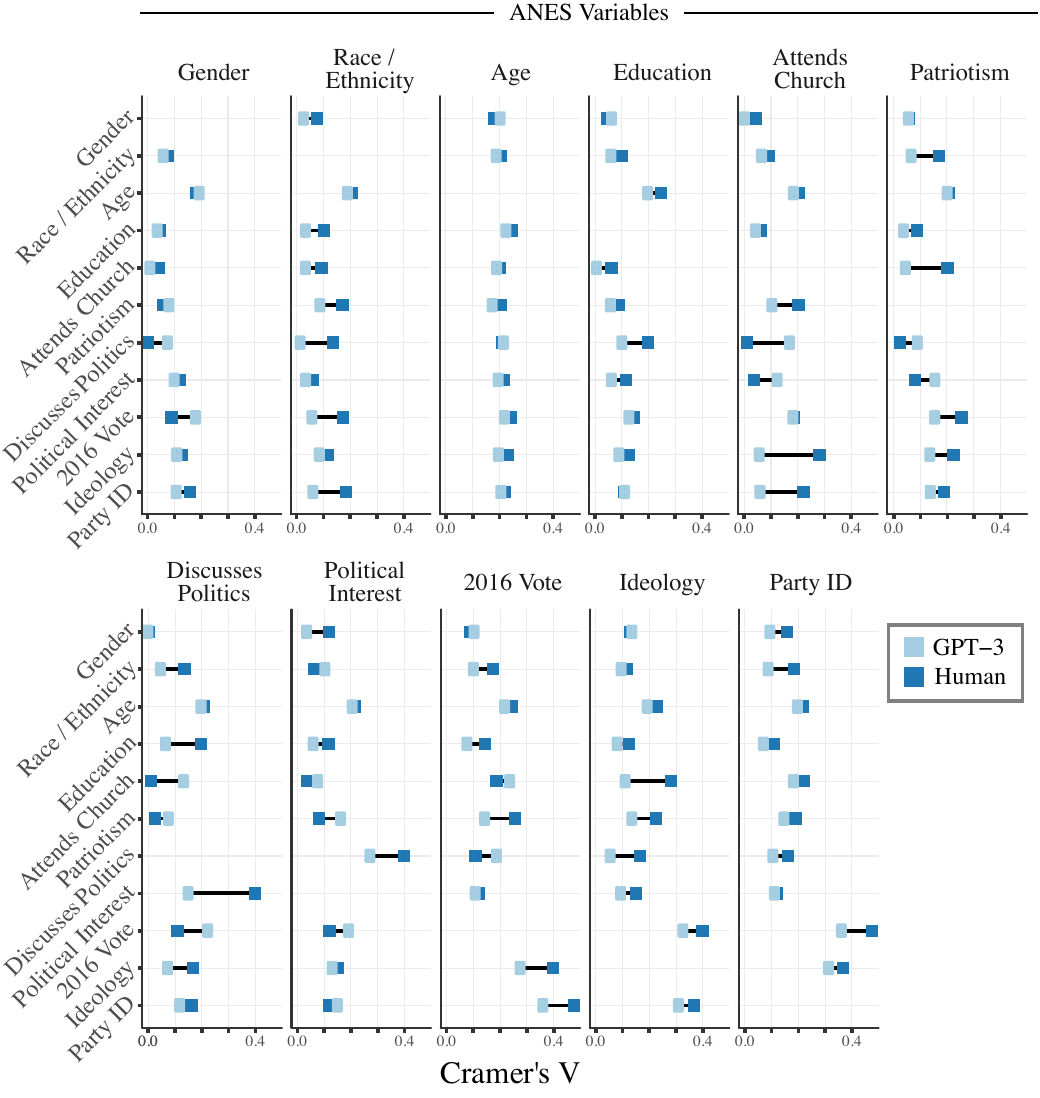}
\caption{Cramer's V Correlations in ANES vs. GPT-3 Data}
\label{fig:anesgpt3}
\end{figure*}

Using the ANES and silicon data, we calculate Cramer's V for each combination of survey items in the ANES sample (``Human''), and between the ANES conditioning values and the resulting GPT-3 produced answer (``GPT-3''). Cramer's V provides a simple summary measure of association that accounts for the variation in base rates in the raw data \citep{Cramer1946}. Figure \ref{fig:anesgpt3} displays the comparison in Cramer's V between the two data sources.
We again find remarkably high correspondence between the patterns of associations in human survey data and these same patterns in GPT-3 produced survey data. The mean difference between the Cramer's V values is -0.026. As can be seen, the Cramer's V for GPT-3-generated responses is not uniformly high or low, but instead mirrors stronger and weaker relationships present in the human data. Where two concepts are not strongly associated in the human data, they likewise show little association in the GPT-3 data. The converse is also true. And while there is variation in Figure \ref{fig:anesgpt3} in terms of how precisely the patterns of relationships in GPT-3 match those in the ANES, the overall pattern is a stunning correspondence between GPT-3 and the ANES in the vast majority of cases. 

Although we provide first-person backstories based on specific human survey profiles, we do not expect the values in the silicon sample to \textit{exactly} match the human response on the individual level. For each text completion, the language model uses a stochastic sampling process to select the completion from the distribution of probable next tokens. Therefore, with a large enough sample size we expect the overall distribution of text responses in the silicon sample to match the overall distribution in the human data, but we do not evaluate correspondence at the individual level. Additionally, as with all stochastic processes, we expect some variation in different draws of the silicon sample. In the appendix, we report on variation in the pattern correspondence based on different sampling parameters in GPT-3.

These results again provide compelling, consistent, repeated evidence for Criterion 4 (Pattern Correspondence). GPT-3 reproduces nuanced patterns of associations not limited to aggregated toplines. When provided with real survey data as inputs, GPT-3 reliably answers closed-ended survey questions in a way that closely mirrors answers given by human respondents. The statistical similarities extend to a whole set of inter-correlations between measures of personal behaviors, demographic characteristics, and complex attitudes. We again see this as strong evidence for algorithmic fidelity.

\section{Where do we go from here?}

Thus far, our emphasis has been on documenting the algorithmic fidelity of GPT-3 by comparing its outputs to human data. However, our purpose is not limited  to these particular human-computer comparisons; if this were the case, the usefulness of GPT-3 would be dramatically limited. Instead, we provide evidence that algorithmic fidelity is a crucial attribute of tools like GPT-3 because it demonstrates that these language models can be used prior to or in the absence of human data.

To illustrate this, consider the conclusions we would have drawn with only our data from GPT-3. The data from our silicon sample in Study 1 suggests that (1) people describe Republicans and Democrats with different terms that highlight distinct stereotypes of both groups; (2) the affective content and extremity of these texts is tied to individuals' political beliefs and identity in systematic ways that can be used to generate theory; (3) stereotypes of partisans contain issue, group, and trait-based content, although trait references are most common; and (4) others can guess the partisanship of individuals based on their stereotypes of Democrats and Republicans. All of this is evident \textit{using only the data from GPT-3}. With this information, interested researchers could design survey questions, experimental treatments, and codebooks to guide human research. Crucially, this can be done with substantially fewer resources than a parallel data collection with human respondents: Study 1 cost \$29 on GPT-3.

The same is true for studies 2 and 3. 
The ablation analysis for Study 2 (contained in Appendix 3) suggests which variables researchers should include in their studies of public opinion if they want to accurately understand Americans' voting behavior. Study 3 could be used by social scientists to target important connections between characteristics and views that merit further exploration. Based on the results from GPT-3, a social scientist could design an experiment or observational study to confirm and dissect this relationship in a rigorous and causal way. The results also indicate which variables operate as potential confounds that should be included in pre-analysis plans for regression and other econometric models that have causal aspirations. Again, all of these insights would be clear to researchers with only access to GPT-3 and without our human baselines. These studies suggest that after establishing algorithmic fidelity in a given model for a given topic/domain, researchers can leverage the insights gained from simulated, silicon samples to pilot different question wording, triage different types of measures, identify key relationships to evaluate more closely, and come up with analysis plans prior to collecting any data with human participants.

%
%

\section{Discussion}

In this paper, we introduce the concept of algorithmic fidelity as a means to justify the use of large-scale language models as proxies for human cognition at an aggregate level, and as general-purpose windows into human thinking. We propose four criteria to establish the algorithmic fidelity of these models and demonstrate empirical methods for their evaluation.

Using these concepts and methods, we show that GPT-3, one of the largest publicly available language models, contains a striking degree of algorithmic fidelity within the realm of public opinion in the United States. Study 1 shows that GPT-3 passes a social science version of the Turing Test (Criterion 1) and exhibits both strong Backward Continuity (Criterion 2) and Pattern Correspondence (Criterion 4). Studies 2 and 3 provide compelling evidence of Forward Continuity (Criterion 3) as well as additional, much more granular evidence for Pattern Correspondence (Criterion 4). As noted in Studies 2 and 3, careful conditioning of GPT-3 allows us to address issues of temporality and replicability, points further supported in results presented in the appendix. Importantly, in all studies, we find evidence that GPT-3 is capable of replicating the viewpoints of demographically varied sub-populations within the U.S. Taken together, these studies show consistent, repeated evidence for these criteria across a range of data sources, measures, and points in time.

These studies also provide examples of some of the many ways in which large scale language models like GPT-3 might be used for social science research. We can envision many others, and expect that this method will have strengths and weaknesses in comparison to traditional social science methods (as we highlight in Appendix 5, cost is certainly a strength of this method). We note, however, that while this work lays exciting groundwork for the beneficial use of these models in social science, these these tools also have dangerous potential. Models with such fidelity, coupled with other computational and methodological advances, could be used to target human groups for misinformation, manipulation, fraud, and so forth \citep{brown2020language}. We acknowledge these dangers, and both join with and strongly endorse the work of others in pushing for a clear standard of ethics for their use in research and deployment \citep{ross2012guide, salganik_bit_2017}. We believe that transparent, research-based, and community-accountable exploration and understanding of these tools will be essential for recognizing and preventing abuse by private actors who will inevitably employ these models for less noble ends.

While the current study is restricted to a specific domain, the underlying methodology is general purpose and calls for additional work to quantify both the extent and limitations of GPT-3's algorithmic fidelity in a wide array of social science fields. Such an effort goes well beyond what one research team can hope to accomplish; we  extend this invitation to the wider scientific community. 



\appendix

\section{General details on GPT-3 usage}
\label{sec:gpt3usage}


For GPT-3 model specifics, refer to Brown et. al.'s original paper from OpenAI referenced in the main text. We use the model through their remote API. This interface accepts several inputs, including a text prompt (e.g. ``backstories'', survey questions, etc.), model specification (we use Davinci, the largest of the models at 175 billion parameters, as opposed to Ada, Curie, or Babbage), and temperature (we use 0.7), and returns a dictionary including text completion and corresponding log-probabilities.

\begin{figure*}
\includegraphics[width=1\linewidth]{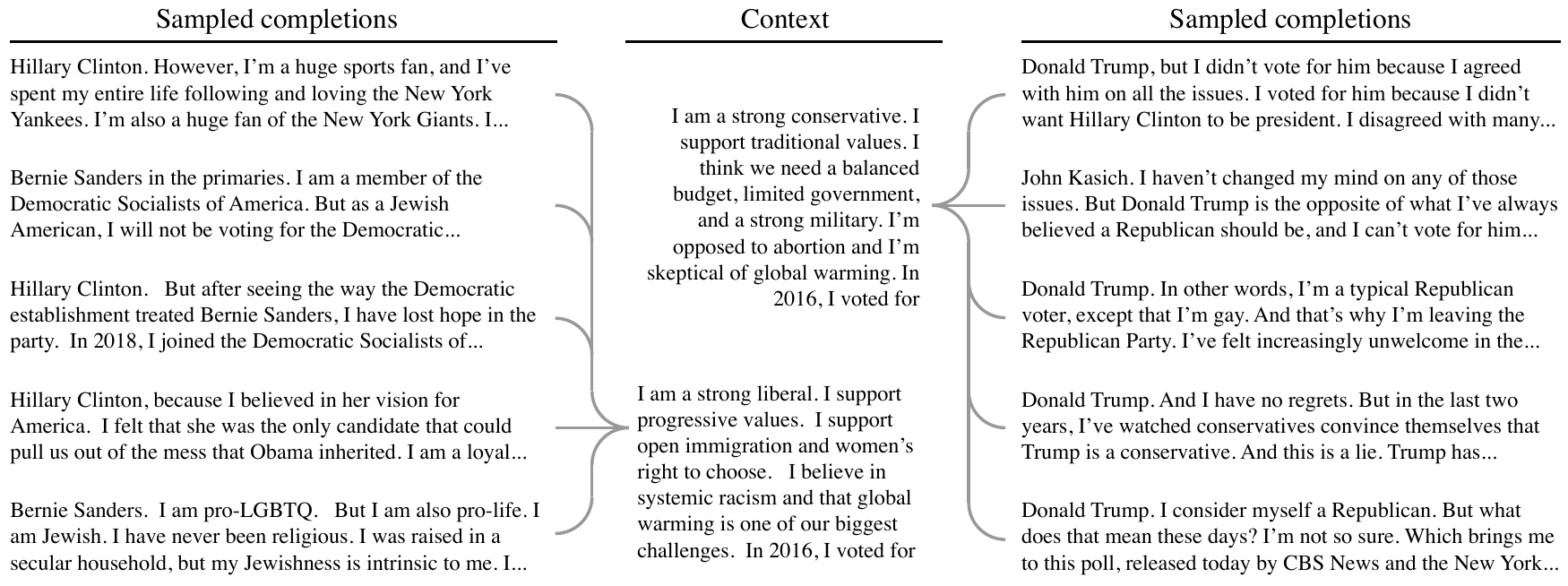}
\caption{Conditioning GPT-3 on first-person demographic backstories yields plausible voting patterns and additional simulated beliefs and opinions.}
\label{fig:conditioning}
\end{figure*}

In contexts where we care about modeling probabilities, as opposed to simply sampling to generate text, we use a standard log-sum-exp conversion. In such contexts, we consider certain token sets to be equivalent. For example, when estimating the probability that a voter cast a vote for Donald Trump in the 2016 presidential election, the prompt might be ``In 2016, I voted for''. Each token in the set \{Donald, donald, DONALD, Trump, trump, TRUMP\} has a distinct share of probability, but we consider them to be different expressions of the same idea. Thus, we sum their probabilities to estimate the un-normalized total share of probability that Donald Trump is being referred to following the prompt. After collapsing token sets and summing their probabilities, we normalize across the remaining collapsed token sets such that their probabilities sum to 1.

In Studies 1 and 3, where we are primarily concerned with text generation, we use the Completion API with default settings, except that temperature is set to 0.7.  This encourages more deterministic answers, at the expense of response diversity.  This parameter was not tuned; it would be interesting to see how our results might vary as it changes.

Figure \ref{fig:conditioning} illustrates some conditioning contexts comprised of first-person demographic backstories paired with multiple sampled continuations generated by GPT-3.






\section{Details on Study 1}

\subsection{Details on Human and GPT-3 samples}

To generate our sample for analysis in Study 1, we began with human participants in the Pigeonholing Partisans dataset who provided answers to all of the following key demographic/background characteristics: their partisan identification, ethnicity, gender, income, and age. These characteristics were identified by the authors of the Pigeonholing Partisans paper to correlate with evaluations of partisan outgroups. 2,107 individuals answered all questions and thus were included in our analysis.

\begin{figure}
\includegraphics[width=1\linewidth]{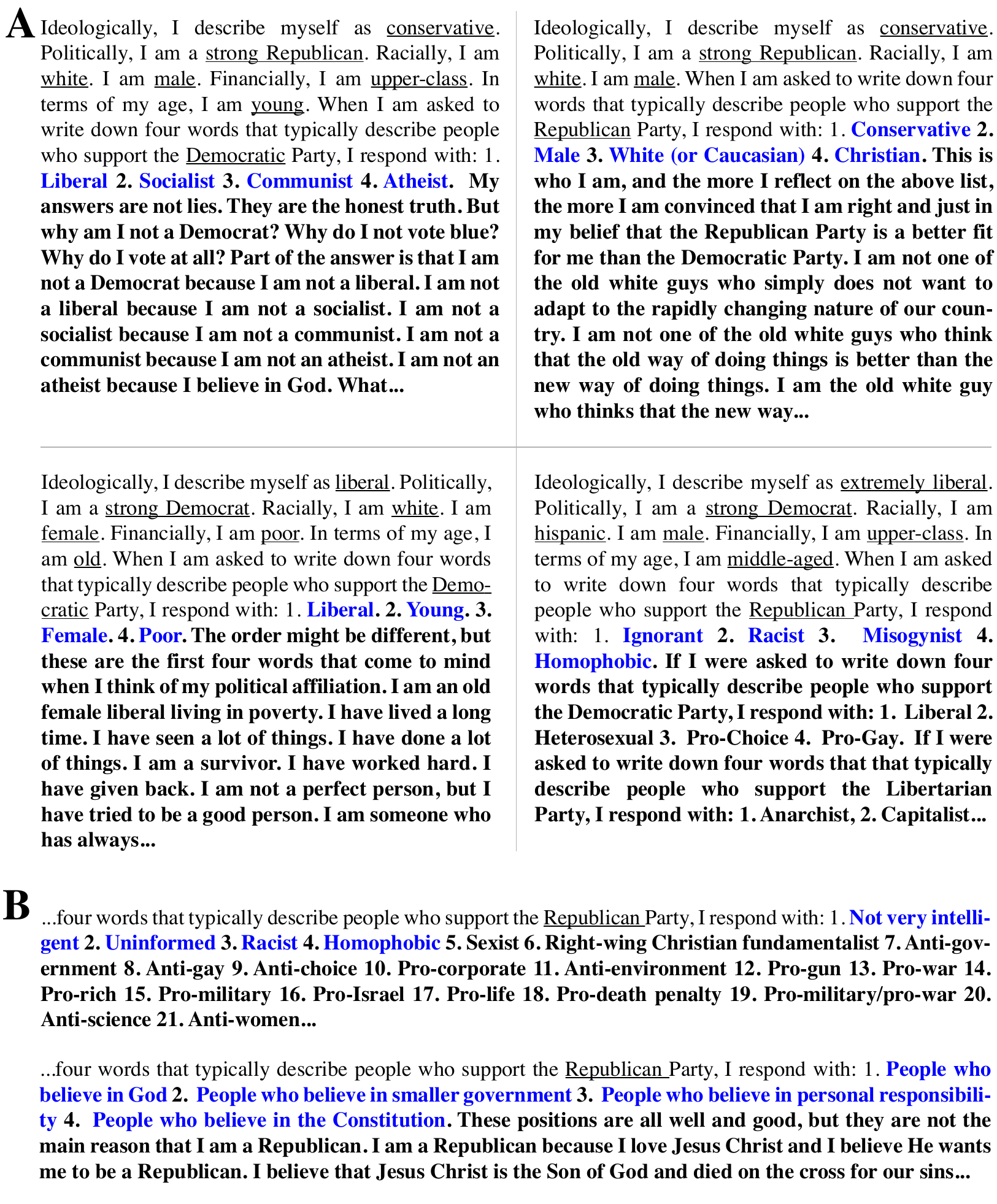}
\caption{Panel A and B: expanded version of Figure 2 in the main text.  Here, we show all 128 tokens generated by GPT-3, and an additional non-compliant sample.}
\label{fig:pp_context_full}
\end{figure}

We then generated a matching set of 2,107 ``silicon'' individuals in GPT-3 by feeding these background characteristics, individual by individual, into a conditioning text template as illustrated in Figure \ref{fig:pp_context_full}. The underlined text in the figure indicates where we plugged each characteristic into GPT-3.
Backstories always concatenated template fragments in the same order (ideology, 7-point partisanship, race, gender, income, age).  If any demographic variable was missing, the corresponding template fragment was omitted.
The age template fragment mapped 18-24 years to the phrase ``young'', 25-39 to ``middle-aged'', 40-60 to ``old'', and 61+ to ``very old''.  The income template fragment mapped annual income less than \$15k to ``very poor'', \$15k-\$50k to ``poor'', \$50k-\$150k to ``middle-class'', and \$150k+ to ``upper-class''.  Other template fragments are self-explanatory.

Using the OpenAI GPT-3 API, we generated 128 tokens worth of text from each silicon respondent. Figure  \ref{fig:pp_context_full} expands on Figure 2 in the text to illustrate what the full responses looked like, with GPT-3 generated text listed in bold. We used regular expressions to extract the four-words at the center of our study.  Light manual post-processing was used to correct situations where the regular expressions were insufficient to extract responses.  If a GPT-3 response listed more than four words or phrases, only the first four were used.  If a GPT-3 response listed less than four, the remaining phrases were left blank.

Both human and GPT-3 ``subjects" were asked to write two lists of words: one describing Republicans, and one describing Democrats. If all participants fully complied, this would mean a total of 2,107 x 2 = 4,214 texts from each sample. As is common in human studies, we didn't receive full compliance: some participants refused to write either list, some only wrote one or the other, and some wrote paragraphs that could not be broken into four categories. After culling out these non-compliant responses, we ended with 3,592 total texts from the human sample (an average of 1.7 texts per respondent), and 4,083 from GPT-3 (1.9 per respondent). GPT-3 was more compliant at this stage of the process. In total, this made 7675 unique lists for analysis.

As can be seen, GPT-3 (like some of our human respondents), sometimes listed more than four words. The most common ``non-compliant" response from GPT-3 was to provide four descriptions, rather than just four words, as illustrated in Panel B. Some of our human respondents did the same. We included all of these descriptive phrase responses in our dataset. As such, some of our study participants saw four phrases, instead of four words.

\begin{figure}
\includegraphics[width=1\linewidth]{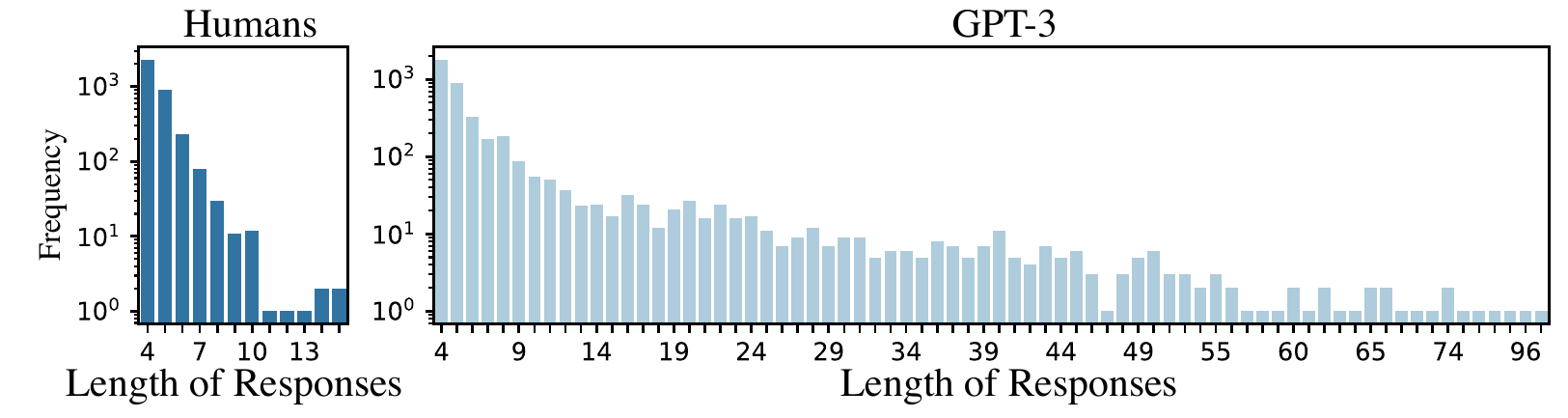}
\caption{Word length of responses in Pigeonholing Partisans data used in the Lucid experiment}
\label{fig:wl}
\end{figure}

As Figure \ref{fig:wl} indicates, Human and GPT-3 respondents differed in their degree of compliance in listing just four words, with GPT-3 including more responses of additional length (note the log scale of the y-axis). The mean human text was 4.54 words long (min = 4; max = 15). The mean GPT-3 text length was 7.78 (min = 4; max = 97). Overall, compliance was high: the modal response in both was 4, and most of the longer responses were 2-3 word phrases in place of single words.

\subsection{Lucid survey design}


We built a survey through \textit{Qualtrics} as an  instrument for these texts to be evaluated, and hired 2,873 individuals from Lucid to do the evaluating. Hiring evaluators from Lucid was faster, cheaper, and provided us with a wider range of types of evaluators than if we had followed the more traditional process of training a small set of research assistants. We designed the survey such that it asked participant to make content-based judgments about 8 randomly-assigned lists from our corpus of 7,675, and then make Turing-test judgments about 8 additional randomly-assigned lists. No respondents evaluated the same texts in both parts of the survey. By design, each text was to be evaluated approximately three times in the content portion and three times in the Turing test portion of the survey. However, due to very minimal non-response, and a few minutes of issues with our server interacting with Lucid's server at the onset of the survey, 120/7,675 (1.5\%) texts were coded only twice, and 7 were coded 4 times. Results do not differ when we exclude these texts.

Lucid participants saw the online/Qualtrics version of the following survey. The order of the answer choices in each of the following questions was randomized across respondents, but kept constant within each respondent. We include a bolded label to highlight each part of the survey in the reproduction of our survey that follows; this label was not shown to respondents: \\

\subsubsection*{Welcome Screen}
Thanks for participating in this project. We expect this task to take 10-12 minutes to complete. If you have not completed it in 1 hour after you begin, the task will expire.

In what follows, you will see 8 short lists of words written to describe Republicans and Democrats. In rare cases, you may see words that are vulgar, offensive, or nonsensical.

On the first page for each list, you will be asked to:
\begin{itemize}
    \item First, guess if the person who wrote the list was themself a Republican, Democrat, or an Independent.
    \item Second, indicate if you feel the words in the list as a whole are positive or negative.
    \item Third, indicate if you think the words listed are extreme.
\end{itemize}

On the next page for each list you will be asked to indicate whether the list of words does or does not include any mention to each of the following:
\begin{enumerate}
\item personality or character traits;
\item specific government or policy issues; or
\item social groups
\end{enumerate}

You’ll then be asked some concluding questions. Please click on the button below to begin. \\

\noindent\textbf{Participants next completed a CAPTCHA item to help prevent bots from completing the instrument} \\

\subsubsection*{List Evaluation Task, shown consecutively for 8 different lists}
Consider the following description of [\textit{Republicans/Democrats; text piped in here}]:
\begin{enumerate}
    \item (\textit{word/phrase 1 piped in here})
    \item (\textit{word/phrase 2 piped in here})
    \item (\textit{word/phrase 3 piped in here})
    \item (\textit{word/phrase 4 piped in here})
\end{enumerate}

\noindent\textbf{Party} Would you say that the person that wrote these words is a Republican, Independent, or Democrat?
\begin{itemize}
  \item Republican
  \item Independent
  \item Democrat
\end{itemize}
\textbf{Positivity} Would you say that this set of words, as a whole, is more positive or more negative?
\begin{itemize}
  \item Very positive
  \item A little positive
  \item Neither positive nor negative
  \item A  little negative
  \item Very negative
\end{itemize}
\textbf{Extremity} Is this set of words extreme?
\begin{itemize}
  \item Yes
  \item No
\end{itemize}
\textbf{Traits} Do these words mention personality or character traits?
\begin{itemize}
  \item Yes
  \item No
\end{itemize}
\textbf{Issues} Do these words include government or policy issues?
\begin{itemize}
  \item Yes
  \item No
\end{itemize}
\textbf{Groups} Do these words mention social groups?
\begin{itemize}
  \item Yes
  \item No
\end{itemize}

\noindent\textbf{After answering these questions for 8 randomly assigned lists, individuals then moved to the Turing task portion of the survey}

\subsubsection*{Turing task introduction screen}
Now, please look at 8 more short sets of words about Republicans and Democrats. Some of these responses were written by people and others were created by a computer program. You may see a few responses from a computer and a few from a person. Or you may see mostly responses from one or another. We want you to guess if a response came from a person or from a computer.

\noindent Please click on the button below to begin.

\subsubsection*{Turing Evaluation Task, shown consecutively for 8 different lists}

Consider the following description of [\textit{Republicans/Democrats; text piped in here}]:
\begin{enumerate}
    \item (\textit{word/phrase 1 piped in here})
    \item (\textit{word/phrase 2 piped in here})
    \item (\textit{word/phrase 3 piped in here})
    \item (\textit{word/phrase 4 piped in here})
\end{enumerate}

\noindent\textbf{Turing task} Would you say that this set of words about [\textit{Republicans/Democrats; text piped in here}] was created by a person or a computer program?
\begin{itemize}
  \item Person
  \item Computer program
\end{itemize}

\noindent\textbf{This same question was used to evaluate all 8 lists}

\subsubsection*{Comments screen}
We appreciate your participation in this survey. If you have any comments, feel free to leave them in the space below.

\subsubsection*{Results screen}
As part of this survey, we asked you to judge if a set of words was created by a person or a computer program. You may be interested in how well you guessed - the table below shows the set of words, your guess, and if the response came from a computer program or a person.

\noindent\textit{Coders were then shown a table with the texts, their guesses, and the correct answers.}

\subsection{Lucid results analysis}

As described in the text, we estimated regression models using Ordinary Least Squares (OLS) to analyze our results. Given that the dependent variable in many of our models is binary (0/1), this means many of these models are linear probability models (LPMs). Results do not significantly differ when we estimate the LPM results using logit instead. As noted in the main text, all models include fixed effects for study participants (recall that each evaluated 8 lists), and clustered standard errors by participants and list (as each list was evaluated three times). We estimated all of these models using the \textit{fixest} R package.

In addition to a binary variable indicting the source of the text, all models include a standard set of variables to control for the potential impact of characteristics of the original list writers on our outcomes. These characteristics come from the original Pigeonholing Partisans dataset, and include the list-writers':

\begin{itemize}
    \item Gender: a categorical variable coded Male, Female, or Other
    \item Ethnicity: two binary categorical variables, Hispanic/Not-Hispanic and White/Other. We include both as controls in our models
    \item Income: Originally asked on an 11-point scale (1 = ``Less than \$15K", 11 = ``More than \$1,000K"). We collapsed this scale to run 0 to 1.
    \item Age: a numeric variable capturing each participant's age, and
    \item Party Identification: a categorical variable coded Democrat, Republican, or Independent.
\end{itemize}

In the ``Percent correctly predicted" model, we add one additional control, for word length (coded numerically as the number of words in each list).

In the main text, we graphically present predicted values from these models. Here we present the full tables of results behind those predictions. As predicted values can only be generated using defined  levels for each of the variables in the model, we chose the following levels: Female, Not Hispanic, White, mean income, mean age, and Democrat. These were the same across all models that included these variables. In the `Percent correctly predicted" model, we set the word length variable to its mean.

Table \ref{tab:codcontent} presents the full results of the models used to predict the percent of texts evaluated as having each of the five characteristics described in the study. These results are presented graphically in Panel B of Figure 4 in the paper.

Table \ref{tab:codcorr} presents the full results of the model used to predict the percent of texts for which Lucid participants correctly guessed the partisanship of the text writer (the top-left bars in Panel B of Figure 4 in the paper).

Tables \ref{tab:f4a1}-\ref{tab:f4a4} present the full results of the models used to generate the predictions in panel A of Figure 4 in the paper. Table \ref{tab:f4a1} corresponds to the top left figure in panel A, Table \ref{tab:f4a2} to the top right, Table \ref{tab:f4a3} to the bottom left, and Table \ref{tab:f4a4} to the bottom right. These models were subset by the ideology of the list writers (using the standard 7-point scale described in the paper). In the tables: EC = Extremely Conservative, C = Conservative, SC = Slightly Conservative, I = Independent, SL = Slightly Liberal, L = Liberal, and EL = Extremely Liberal.

\begin{table}
\centering
\begin{tabular}{lccccc}
\toprule
&Positive&Extreme&Traits&Issues&Groups\\
\midrule Source:GPT-3&-0.010 & 0.013 & -0.058 & 0.033 & 0.078\\
  &(0.006) & (0.007) & (0.007) & (0.007) & (0.007)\\
Gender:Female&-0.002 & -0.006 & -0.010 & 0.013 & 0.009\\
  &(0.006) & (0.008) & (0.007) & (0.007) & (0.007)\\
Gender:Other&-0.111 & 0.129 & -0.175 & 0.036 & -0.077\\
  &(0.048) & (0.073) & (0.072) & (0.063) & (0.042)\\
Not Hispanic&0.019 & -0.011 & 0.003 & -0.002 & -0.0002\\
  &(0.009) & (0.012) & (0.011) & (0.012) & (0.012)\\
Income&0.009 & -0.008 & -0.003 & 0.007 & 0.003\\
  &(0.001) & (0.002) & (0.002) & (0.002) & (0.002)\\
White&0.001 & -0.002 & 0.021 & 0.017 & -0.011\\
  &(0.007) & (0.009) & (0.008) & (0.008) & (0.008)\\
Age&-0.0005 & 0.001 & 0.0003 & 0.00006 & -0.0003\\
  &(0.0002) & (0.0002) & (0.0002) & (0.0002) & (0.0002)\\
PID:Indep.&-0.029 & 0.031 & -0.005 & 0.018 & -0.010\\
  &(0.009) & (0.012) & (0.011) & (0.011) & (0.011)\\
PID:Rep.&0.011 & -0.022 & -0.034 & 0.027 & -0.012\\
  &(0.007) & (0.009) & (0.008) & (0.008) & (0.008)\\
 &  & & & & \\
Observations & 18,903&18,903&18,903&18,903&18,903\\
RMSE & 0.28971&0.39470&0.36560&0.36634&0.37094\\
 &   &   &   &   &  \\
Evaluators fixed effects & $\checkmark$ & $\checkmark$ & $\checkmark$ & $\checkmark$ & $\checkmark$\\
\bottomrule
\multicolumn{6}{l}{\emph{Two-way (Evaluators \& Lists) standard-errors in parentheses}}\\
\end{tabular}
\caption{Evaluated content of lists, Lucid Experiment}
\label{tab:codcontent}
\end{table}

\begin{table}
\centering
\begin{tabular}{lc}
\toprule
&Percent Correctly Guessed\\
\midrule Source:GPT-3&-0.073\\
  &(0.008)\\
Gender:Female&-0.007\\
  &(0.008)\\
Gender:Other&-0.059\\
  &(0.082)\\
Not Hispanic&-0.006\\
  &(0.013)\\
Income&0.002\\
  &(0.002)\\
White&0.012\\
  &(0.010)\\
Age&0.0007\\
  &(0.0003)\\
Word Length&0.0009\\
  &(0.0007)\\
PID:Indep.&-0.285\\
  &(0.014)\\
PID:Rep.&-0.097\\
  &(0.011)\\
 &  \\
Observations & 18,903\\
RMSE & 0.43912\\
 &  \\
Evaluators fixed effects & $\checkmark$\\
\bottomrule
\multicolumn{2}{l}{\emph{Two-way (Evaluators \& Lists) standard-errors in parentheses}}\\
\end{tabular}
\caption{Correctly guessing the partisanship of list writers, Lucid Experiment}
\label{tab:codcorr}
\end{table}

\begin{table}
\centering
\begin{tabular}{lccccccc}
\toprule
&\multicolumn{7}{c}{Positive}\\
& EC & C & SC & I & SL & L & EL\\
\midrule Source:GPT-3&-0.085 & 0.019 & -0.041 & 0.040 & 0.018 & 0.034 & 0.019\\
  &(0.121) & (0.046) & (0.054) & (0.028) & (0.033) & (0.019) & (0.051)\\
Gender:Female&0.015 & -0.004 & -0.079 & 0.009 & -0.034 & -0.055 & 0.072\\
  &(0.103) & (0.047) & (0.050) & (0.028) & (0.037) & (0.019) & (0.041)\\
Not Hispanic&0.055 & -0.102 & 0.040 & 0.103 & 0.023 & -0.032 & 0.022\\
  &(0.211) & (0.095) & (0.064) & (0.043) & (0.056) & (0.037) & (0.037)\\
Income&0.011 & 0.002 & -0.002 & 0.011 & 0.012 & 0.005 & 0.017\\
  &(0.025) & (0.010) & (0.009) & (0.007) & (0.007) & (0.004) & (0.009)\\
White&-0.013 & 0.147 & 0.035 & 0.075 & 0.033 & 0.009 & -0.086\\
  &(0.157) & (0.078) & (0.058) & (0.034) & (0.041) & (0.024) & (0.053)\\
Age&-0.012 & -0.001 & -0.002 & -0.003 & -0.0001 & -0.0010 & -0.002\\
  &(0.005) & (0.001) & (0.001) & (0.0009) & (0.001) & (0.0006) & (0.001)\\
Gender:Other&   &    &    & -0.487 &    &    & -0.078\\
  &   &    &    & (0.163) &    &    & (0.056)\\
 &  & & & & & & \\
Observations & 387&1,122&1,059&2,072&1,419&2,374&1,036\\
RMSE & 0.05609&0.11621&0.10735&0.15728&0.12102&0.13797&0.08714\\
 &   &   &   &   &   &   &  \\
Evaluators fixed effects & $\checkmark$ & $\checkmark$ & $\checkmark$ & $\checkmark$ & $\checkmark$ & $\checkmark$ & $\checkmark$\\
\bottomrule
\multicolumn{8}{l}{\emph{Two-way (Evaluators \& Lists) standard-errors in parentheses}}\\
\end{tabular}
\caption{Percent of texts rated positively, subset by the ideology of individual list writers: Describing Republicans}
\label{tab:f4a1}
\end{table}

\begin{table}
\centering
\begin{tabular}{lccccccc}
\toprule
&\multicolumn{7}{c}{Positive}\\
& EC & C & SC & I & SL & L & EL\\
\midrule Source:GPT-3&-0.145 & 0.044 & 0.041 & 0.089 & -0.028 & -0.099 & -0.033\\
  &(0.172) & (0.044) & (0.045) & (0.027) & (0.032) & (0.020) & (0.053)\\
Gender:Female&-0.198 & 0.011 & 0.023 & -0.026 & -0.013 & 0.003 & 0.081\\
  &(0.119) & (0.057) & (0.042) & (0.026) & (0.034) & (0.019) & (0.045)\\
Not Hispanic&0.568 & 0.112 & -0.093 & 0.0004 & -0.084 & 0.051 & 0.089\\
  &(0.226) & (0.111) & (0.089) & (0.039) & (0.051) & (0.037) & (0.068)\\
Income&-0.037 & 0.006 & 0.005 & 0.010 & 0.020 & 0.011 & 0.040\\
  &(0.052) & (0.010) & (0.010) & (0.007) & (0.008) & (0.004) & (0.011)\\
White&0.917 & -0.006 & -0.077 & -0.015 & -0.046 & 0.038 & -0.047\\
  &(0.177) & (0.067) & (0.043) & (0.031) & (0.043) & (0.024) & (0.067)\\
Age&-0.006 & -0.002 & -0.0003 & 0.0002 & 0.001 & -0.001 & -0.0004\\
  &(0.002) & (0.001) & (0.001) & (0.0009) & (0.001) & (0.0007) & (0.002)\\
Gender:Other&   &    &    & -0.014 &    &    & 0.190\\
  &   &    &    & (0.061) &    &    & (0.109)\\
 &  & & & & & & \\
Observations & 393&1,121&1,048&2,062&1,423&2,370&1,029\\
RMSE & 0.04445&0.11032&0.10969&0.15234&0.11873&0.13810&0.09856\\
 &   &   &   &   &   &   &  \\
Evaluators fixed effects & $\checkmark$ & $\checkmark$ & $\checkmark$ & $\checkmark$ & $\checkmark$ & $\checkmark$ & $\checkmark$\\
\bottomrule
\multicolumn{8}{l}{\emph{Two-way (Evaluators \& Lists) standard-errors in parentheses}}\\
\end{tabular}
\caption{Percent of texts rated positively, subset by the ideology of individual list writers: Describing Democrats}
\label{tab:f4a2}
\end{table}

\begin{table}
\centering
\begin{tabular}{lccccccc}
\toprule
&\multicolumn{7}{c}{Extreme}\\
& EC & C & SC & I & SL & L & EL\\
\midrule Source:GPT-3&0.079 & -0.025 & 0.064 & 0.035 & -0.066 & 0.016 & 0.143\\
  &(0.070) & (0.060) & (0.069) & (0.038) & (0.051) & (0.034) & (0.080)\\
Gender:Female&0.024 & 0.061 & -0.137 & -0.035 & 0.059 & 0.036 & 0.054\\
  &(0.027) & (0.059) & (0.067) & (0.037) & (0.056) & (0.034) & (0.065)\\
Not Hispanic&0.023 & -0.048 & -0.075 & -0.093 & -0.024 & -0.029 & -0.108\\
  &(0.028) & (0.119) & (0.103) & (0.066) & (0.095) & (0.061) & (0.083)\\
Income&0.005 & 0.023 & 0.023 & -0.018 & 0.006 & -0.008 & 0.012\\
  &(0.009) & (0.013) & (0.013) & (0.009) & (0.012) & (0.008) & (0.017)\\
White&-0.097 & -0.241 & -0.009 & -0.137 & 0.046 & -0.010 & -0.073\\
  &(0.093) & (0.107) & (0.075) & (0.047) & (0.067) & (0.038) & (0.094)\\
Age&0.004 & 0.0010 & 0.002 & 0.004 & 0.001 & 0.0009 & 0.008\\
  &(0.003) & (0.002) & (0.002) & (0.001) & (0.002) & (0.001) & (0.002)\\
Gender:Other&   &    &    & 0.705 &    &    & 0.435\\
  &   &    &    & (0.273) &    &    & (0.308)\\
 &  & & & & & & \\
Observations & 387&1,122&1,059&2,072&1,419&2,374&1,036\\
RMSE & 0.03172&0.15942&0.15086&0.21933&0.18450&0.23722&0.15852\\
 &   &   &   &   &   &   &  \\
Evaluators fixed effects & $\checkmark$ & $\checkmark$ & $\checkmark$ & $\checkmark$ & $\checkmark$ & $\checkmark$ & $\checkmark$\\
\bottomrule
\multicolumn{8}{l}{\emph{Two-way (Evaluators \& Lists) standard-errors in parentheses}}\\
\end{tabular}
\caption{Percent of texts rated as extreme, subset by the ideology of individual list writers: Describing Republicans}
\label{tab:f4a3}
\end{table}

\begin{table}
\centering
\begin{tabular}{lccccccc}
\toprule
&\multicolumn{7}{c}{Extreme}\\
& EC & C & SC & I & SL & L & EL\\
\midrule Source:GPT-3&-0.374 & -0.089 & -0.024 & -0.062 & 0.027 & 0.015 & -0.030\\
  &(0.277) & (0.059) & (0.057) & (0.035) & (0.044) & (0.027) & (0.084)\\
Gender:Female&-0.101 & -0.105 & -0.171 & 0.058 & 0.067 & -0.004 & -0.110\\
  &(0.247) & (0.071) & (0.065) & (0.034) & (0.051) & (0.027) & (0.064)\\
Not Hispanic&0.085 & -0.033 & 0.173 & 0.089 & 0.063 & -0.006 & 0.015\\
  &(0.524) & (0.120) & (0.113) & (0.048) & (0.061) & (0.053) & (0.103)\\
Income&0.002 & -0.009 & -0.005 & -0.019 & -0.003 & -0.002 & -0.005\\
  &(0.062) & (0.014) & (0.015) & (0.008) & (0.011) & (0.006) & (0.017)\\
White&0.462 & 0.103 & -0.073 & -0.025 & 0.013 & 0.005 & 0.053\\
  &(0.317) & (0.091) & (0.068) & (0.038) & (0.068) & (0.032) & (0.103)\\
Age&-0.006 & 0.0008 & 0.003 & -0.0002 & -0.0001 & 0.002 & 0.004\\
  &(0.007) & (0.002) & (0.002) & (0.001) & (0.002) & (0.0009) & (0.003)\\
Gender:Other&   &    &    & 0.146 &    &    & 0.123\\
  &   &    &    & (0.086) &    &    & (0.274)\\
 &  & & & & & & \\
Observations & 393&1,121&1,048&2,062&1,423&2,370&1,029\\
RMSE & 0.06764&0.15862&0.14552&0.20705&0.16029&0.19170&0.14525\\
 &   &   &   &   &   &   &  \\
Evaluators fixed effects & $\checkmark$ & $\checkmark$ & $\checkmark$ & $\checkmark$ & $\checkmark$ & $\checkmark$ & $\checkmark$\\
\bottomrule
\multicolumn{8}{l}{\emph{Two-way (Evaluators \& Lists) standard-errors in parentheses}}\\
\multicolumn{8}{l}{\emph{Signif. Codes: ***: 0.001, **: 0.01, *: 0.05}}\\
\end{tabular}
\caption{Percent of texts rated as extreme, subset by the ideology of individual list writers: Describing Democrats}
\label{tab:f4a4}
\end{table}


\section{Details on Study 2}

\subsection{Data generation}

For study 2, we generated a silicon sample based on the 2012, 2016, and 2020 ANES Timeseries datasets.  For each subject, we constructed a first-person backstory using a templating strategy similar to that in Study 1.  We used the following variables from the ANES to condition GPT-3; in this list, variable names from the datasets follow in parentheses in this order - 2012 / 2016/ 2020. The variables were (1) racial/ethnic self-identification (dem-raceeth-x / V161310x / V201549x), (2) gender (gender$\_$respondent$\_$x / V161342 / V201600), (3) age (dem$\_$age$\_$r$\_$x / V161267 / V201507x), (4) conservative-liberal ideological self-placement (libcpre$\_$self / V161126 / V201200), (5) party identification (pid$\_$x / V161158x / V201231x), (6) if the subject is interested in politics (paprofile$\_$interestpolit / V162256 / V202406), (7) if the respondent attends church (relig$\_$church / V161244 / V201452), (8) if the respondent reported discussing politics with family and friends (discuss$\_$disc / V162174 / V201452), (9) feelings of patriotism associated with the American flag (patriot$\_$flag / V162125x / \textit{Not asked}), and (10) respondents' state of residence (sample$\_$stfips / V161010d / \textit{Not released as of the time of this writing}). For the measure of self-reported vote from the ANES, we used presvote2012$\_$x in 2012, V162062x in 2016, and V202110x in 2020.

For all template fragments, phrasing was selected to closely match the ANES, although the ANES phrasing was translated into first-person declarations.  For the age and state of residence fragments, the ANES result was inserted directly into the template.  All other template fragments mapped the ANES variable to a short string, such as ``attend church'', ``extremely liberal'', ``native American'', etc. that closely matches the corresponding ANES value, which was then inserted into the template fragment.  Template fragments were then concatenated together to create a final backstory. If any variable for any subject was missing, the corresponding template fragment was omitted.

Because this study predicts voting patterns, we are interested in the probability that GPT-3 assigns to voting for a particular candidate, given a specific backstory.
Note that in this study, GPT-3 was not required to sample any completions; we only use it to compute the probability of a single successor token, given the conditioning context.  For this reason, the temperature parameter and sampling strategy of the OpenAI API are irrelevant.
Because GPT-3 assigns some probability to a wide variety of semantically equivalent phrases, we collapse them as described in Section~\ref{sec:gpt3usage}.  We used two token sets for each year of data. In 2012, we included the following token sets for voting for Romney: ``romney'', ``mitt'', ``republican'', and ``conservative''. The token set for voting for Obama was ``obama'', ``barack'', ``democrat'', ``democratic'', and ``liberal''. For 2016, the Trump token set included the terms ``trump'', ``donald'', ``republican'' and ``conservative''. For the 2016 Clinton token set, we included ``clinton'', ``hillary'', ``rodham'', ``senator'', ``democrat'', ``democratic'', and ``liberal''. For the 2020 data,  the token set for Trump included ``trump'', ``donald'', ``republican'' and ``conservative''. For Biden, the token set was ``joe'', ``joseph'', ``biden'', ``democratic'', ``democrat'', and ``liberal''.
For all of these token sets, lexical variations of each term (lower-case, upper-case, mixed-case, with and without leading and trailing spaces, etc.; these are all considered distinct tokens by GPT-3) were also included. Any tokens not in the token sets were ignored.  Token sets were selected to ensure that common cases were caught, but were not tuned or optimized to improve results.

\subsection{Data analysis}

The primary analysis of this silicon sample comes from comparing the vote choice as reported by ANES respondents and the probability for voting for the Republican candidate from GPT-3. To make the predictions from GPT-3 match the observed human data (our baseline in this case), we dichotomized the probability of voting for the Republican candidate from GPT-3 by dividing the responses exactly at 0.50; probabilities of more than .50 were coded as a vote for the Republican (i.e., Romney or Trump) and probabilities lower than 0.50 were coded as a vote for the Democrat (i.e., Obama, Clinton, or Biden). No probabilities were predicted to be exactly 0.50.

This gives us two binary variables, with which we estimated 4 statistics. Table 1 in the main text presented only the tetrachoric corelation and proportion agreement, solely for presentational and space purposes. In the tables in this section, we show the entire set of metrics. In the following table, we calculate the correspondence between the vote variable from the ANES and GPT-3 in four different ways, each of which is a way to determine how closely two binary variables correspond. These statistics are as follows (presented in the same order as in the subsequent tables of results):
\begin{itemize}
  \item \textbf{Tetrachoric correlation:} This measure is a way to calculate a correlation between two variables when both are binary but come from an underlying, continuous, normal distribution. It is similar to Pearson's \textbf{\emph{r}} in it's interpretation: values closer to 1 indicate closer correspondence, and values near 0 indicate almost no correspondence. These values were calculated using the tetrachoric command from the \textit{psych} package in R.
  \item \textbf{Cohen's Kappa:} This statistic, sometimes referred to as $\kappa$, calculates the agreement between two variables. It is generally used to compare the agreement of two raters, and here we use it treating the ANES and GPT-3 estimates as the two ratings. Many prefer this measure over the proportion of agreement because $\kappa$ includes a penalty for the amount of agreement that might have occurred due to chance alone. Values of $\kappa$ typically range from 0 to 1, with the same interpretation as tetrachoric correlation and Pearson's \textbf{\emph{r}}. It is theoretically possible to obtain a negative value for $\kappa$; this would indicate worse correspondence between the variables than would occur by chance. The values in Table 1 were calculated using the cohen.kappa command from the \textit{psych} package in R.
  \item \textbf{Intraclass correlation coefficient  or ICC:} Similar to $\kappa$, ICC is commonly used as a measure of agreement between raters or coders. Values closer to 1 indicate stronger agreement, and generally scores higher than 0.75 are considered indicates of strong agreement. It is more flexible and can be used to compare variables of different measurement metrics (e.g., ordinal, continuous, binary, etc.) to one another. Here we present the results for the ICC measures for the binary vote variables, but replacing the GPT-3 binary variable for the underlying probability does not change the ICC measures in meaningful ways. Given that our interest is understanding how both the human and GPT-3 measures compare to one another, we use the averaged versions of the ICC statistics. Further, rather than focus on a specific measure of ICC (such as ICC1, ICC2, or ICC3), we simply report the \textit{lowest} of the three. In nearly all cases, the differences between these versions of ICC were neglible. Like the previous two statistics, ICC was calculated with the \textit{psych} package in R, specifically with the  ICC command.
  \item \textbf{Proportion agreement:} This is the simplest of the measures and indicates the proportion of the observations where the two vote variables (GPT-3 and human response) exactly match. It does not account for the probability of matching by chance and should be viewed as a descriptive quantity. It was calculated by creating frequency tables of the GPT-3 and ANES vote variables and then calculating proportions based on those frequencies. We include proportion agreement because some of the other measures (such as the tetrachoric correlation and $\kappa$) do not perform well when all of the data (more than 95 percent) fall in the same quadrant of the frequency table. As a concrete example, the correlations and $\kappa$ are quite low for Strong Democrats; upon closer examination, though, this seems to occur because there is almost no variation in the vote variable for GPT-3 or the ANES. There is near complete agreement between both estimates of vote - it is just that all of the respondents reported voting (or are predicted by GPT-3 to vote for) the same candidate. This almost complete lack of variation on the vote variable itself seems to make the measures of correspondence unreliable and unreflective of the agreement between GPT-3 and the ANES.
\end{itemize}

\begin{table}[t!]
\small
\centering
\begin{tabular}{l c c c c }
    \hline
        \thead{Variable} & \thead{Tetrachoric Correlation} & \thead{Cohen’s Kappa} & \thead{ICC} & \thead{Prop. agreement} \\ \hline
        Whole sample & 0.90 & 0.69 & 0.81 & 0.85 \\
        Men & 0.90 & 0.70 & 0.82 & 0.85 \\
        Women & 0.91 & 0.67 & 0.80 & 0.86 \\
        Strong partisans & 0.99 & 0.93 & 0.96 & 0.97 \\
        Weak partisans & 0.73 & 0.45 & 0.61 & 0.74 \\
        Leaners & 0.90 & 0.70 & 0.82 & 0.85 \\
        Independents & 0.31 & 0.16 & 0.22 & 0.59 \\
        Conservatives & 0.84 & 0.59 & 0.74 & 0.84 \\
        Moderates & 0.65 & 0.40 & 0.57 & 0.77 \\
        Liberals & 0.81 & 0.43 & 0.60 & 0.95 \\
        Whites & 0.87 & 0.64 & 0.77 & 0.82 \\
        Blacks & 0.71 & 0.31 & 0.47 & 0.97 \\
        Hispanics & 0.86 & 0.63 & 0.78 & 0.86 \\
        Attends church & 0.91 & 0.71 & 0.83 & 0.86 \\
        Does not attend church & 0.88 & 0.64 & 0.78 & 0.85 \\
        Very interested in politics & 0.95 & 0.80 & 0.89 & 0.90 \\
        Not at all interested in politics & 0.71 & 0.38 & 0.53 & 0.74 \\
        Discusses politics & 0.92 & 0.72 & 0.84 & 0.87 \\
        Does not discuss politics & 0.83 & 0.57 & 0.73 & 0.82 \\
        18 to 30 years old & 0.90 & 0.66 & 0.80 & 0.87 \\
        31 to 45 years old & 0.90 & 0.65 & 0.79 & 0.85 \\
        46 to 60 years old & 0.90 & 0.69 & 0.82 & 0.86 \\
        Over 60 & 0.90 & 0.71 & 0.83 & 0.85 \\
        Californians & 0.92 & 0.62 & 0.76 & 0.85 \\
        Texans & 0.91 & 0.69 & 0.81 & 0.84 \\
        New Yorkers & 0.91 & 0.59 & 0.74 & 0.84 \\
        Ohioans & 0.88 & 0.66 & 0.80 & 0.84 \\
        Arizonans & 0.98 & 0.89 & 0.94 & 0.95 \\
        Wisconsins & 0.95 & 0.70 & 0.82 & 0.85 \\ \hline
    \end{tabular}
    \caption{Various measures of correlation between GPT-3 and ANES probability of voting for Mitt Romney in 2012. GPT-3 vote is a binary version of GPT-3's predicted probability of voting for Mitt Romney, dividing predictions at 0.50.}
\label{tab:anescorr2012}
\end{table}

\begin{table}[t!]
\small
\centering
\begin{tabular}{l c c c c }
    \hline
        \thead{Variable} & \thead{Tetrachoric Correlation} & \thead{Cohen’s Kappa} & \thead{ICC} & \thead{Prop. agreement} \\ \hline
        Whole sample & 0.92 & 0.73 & 0.84 & 0.87 \\
        Men & 0.93 & 0.76 & 0.86 & 0.88 \\
        Women & 0.92 & 0.7 & 0.82 & 0.86 \\
        Strong partisans & 1.00 & 0.95 & 0.97 & 0.97 \\
        Weak partisans & 0.71 & 0.46 & 0.62 & 0.74 \\
        Leaners & 0.93 & 0.74 & 0.85 & 0.87 \\
        Independents & 0.41 & 0.25 & 0.39 & 0.62 \\
        Conservatives & 0.88 & 0.66 & 0.79 & 0.86 \\
        Moderates & 0.76 & 0.52 & 0.69 & 0.78 \\
        Liberals & 0.73 & 0.25 & 0.39 & 0.95 \\
        Whites & 0.91 & 0.7 & 0.83 & 0.85 \\
        Blacks & 0.87 & 0.51 & 0.67 & 0.96 \\
        Hispanics & 0.93 & 0.73 & 0.85 & 0.9 \\
        Attends church & 0.93 & 0.75 & 0.86 & 0.88 \\
        Does not attend church & 0.9 & 0.67 & 0.8 & 0.85 \\
        Very interested in politics & 0.97 & 0.85 & 0.92 & 0.93 \\
        Not at all interested in politics & 0.75 & 0.48 & 0.64 & 0.75 \\
        Discusses politics & 0.94 & 0.76 & 0.86 & 0.88 \\
        Does not discuss politics & 0.81 & 0.57 & 0.72 & 0.79 \\
        18 to 30 years old & 0.9 & 0.69 & 0.81 & 0.86 \\
        31 to 45 years old & 0.92 & 0.72 & 0.84 & 0.87 \\
        46 to 60 years old & 0.92 & 0.72 & 0.83 & 0.86 \\
        Over 60 & 0.93 & 0.75 & 0.85 & 0.87 \\
        Californians & 0.87 & 0.58 & 0.72 & 0.83 \\
        Texans & 0.95 & 0.79 & 0.88 & 0.9 \\
        New Yorkers & 0.95 & 0.79 & 0.89 & 0.91 \\
        Ohioans & 0.9 & 0.7 & 0.83 & 0.85 \\
        Arizonans & 0.92 & 0.74 & 0.85 & 0.87 \\
        Wisconsins & 0.97 & 0.84 & 0.91 & 0.92 \\ \hline
    \end{tabular}
    \caption{Various measures of correlation between GPT-3 and ANES probability of voting for Donald Trump in 2016. GPT-3 vote is a binary version of GPT-3's predicted probability of voting for Donald Trump, dividing predictions at 0.50.}
\label{tab:anescorr2016}
\end{table}

\begin{table}[t!]
\small
\centering
\begin{tabular}{l c c c c }
    \hline
        \thead{Variable} & \thead{Tetrachoric Correlation} & \thead{Cohen’s Kappa} & \thead{ICC} & \thead{Prop. agreement} \\ \hline
        Whole sample & 0.94 & 0.77 & 0.87 & 0.89 \\
        Men & 0.95 & 0.77 & 0.87 & 0.88 \\
        Women & 0.94 & 0.78 & 0.88 & 0.90 \\
        Strong partisans & 1.00 & 0.95 & 0.97 & 0.97 \\
        Weak partisans & 0.84 & 0.63 & 0.77 & 0.82 \\
        Leaners & 0.95 & 0.79 & 0.88 & 0.89 \\
        Independents & 0.02 & 0.02 & 0.03 & 0.53 \\
        Conservatives & 0.91 & 0.71 & 0.83 & 0.89 \\
        Moderates & 0.71 & 0.48 & 0.65 & 0.77 \\
        Liberals & 0.86 & 0.51 & 0.67 & 0.97 \\
        Whites & 0.94 & 0.78 & 0.88 & 0.89 \\
        Blacks & 0.81 & 0.49 & 0.66 & 0.94 \\
        Hispanics & 0.88 & 0.63 & 0.77 & 0.83 \\
        Attends church & 0.94 & 0.77 & 0.87 & 0.88 \\
        Does not attend church & 0.93 & 0.76 & 0.86 & 0.90 \\
        Very interested in politics & 0.97 & 0.84 & 0.91 & 0.92 \\
        Not at all interested in politics & 0.83 & 0.62 & 0.77 & 0.81 \\
        Discusses politics & 0.95 & 0.79 & 0.88 & 0.90 \\
        Does not discuss politics & 0.80 & 0.59 & 0.74 & 0.79 \\
        18 to 30 years old & 0.90 & 0.70 & 0.82 & 0.87 \\
        31 to 45 years old & 0.94 & 0.78 & 0.88 & 0.90 \\
        46 to 60 years old & 0.92 & 0.74 & 0.85 & 0.87 \\
        Over 60 & 0.96 & 0.82 & 0.90 & 0.91 \\
 \hline
    \end{tabular}
    \caption{Various measures of correlation between GPT-3 and ANES probability of voting for Donald Trump in 2020. GPT-3 vote is a binary version of GPT-3's predicted probability of voting for Donald Trump, dividing predictions at 0.50.}
\label{tab:anescorr2020}
\end{table}


\clearpage

\subsection{Ablation analysis}

\begin{figure*}
\includegraphics[width=1\linewidth]{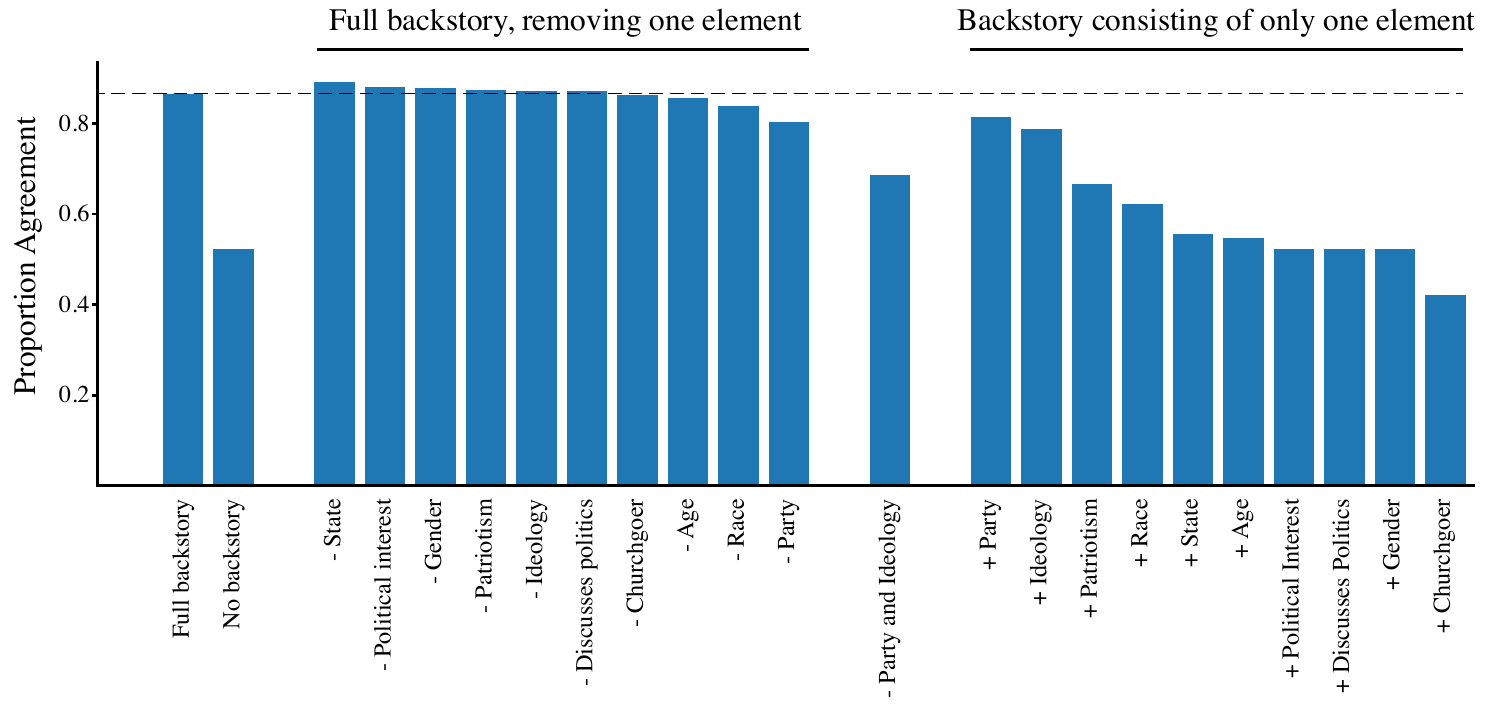}
\caption{An ablation experiment examining the importance of each backstory element. Reported is the Proportion Agreement on the vote prediction task of the ANES 2016 dataset. Each bar represents a different template with some set of backstory elements, from ``Full backstory'' (yielding the results shown in the main paper), to ``No backstory'' (where each silicon subject would have the same, empty backstory, therefore resulting the same vote prediction for every subject; this is essentially equivalent to random chance).}
\label{fig:voting_ablation}
\end{figure*}

We also conducted an ablation study on the backstories used for vote prediction in the ANES 2016 experiment. Recall that each backstory consisted of a template with 10 different elements. For this experiment, we investigated how the elements of the template interacted with each other by systematically removing one or two at a time. We also tested backstories consisting of \emph{only} one backstory element.

The results are shown in Fig.~\ref{fig:voting_ablation}. There are a few notable elements to these results. First, no single backstory element accounted for all of the predicted power of GPT-3's vote predictions, suggesting that GPT-3 is indeed synthesizing or fusing multiple backstory elements together, yielding a more accurate final prediction.  Second, GPT-3 can use either Party or Ideology to predict vote choice, but Party is more predictive. Third, the addition of some elements of the backstory template (such as State or Political Interest) mildly hurt performance.  Finally, we conducted an experiment where we removed both Party and Ideology from the template, yielding only demographic factors; we see that the combination of the remaining 8 elements yields better accuracy than any single element.

We here additionally note that no attempt was made to optimize the template used during our experiments; the template used and the 10 elements selected represent our first try. Future work can likely improve these results by optimizing template and backstory elements before an experiment begins.

\subsection{Model comparison}

\begin{figure*}
\includegraphics[width=1\linewidth]{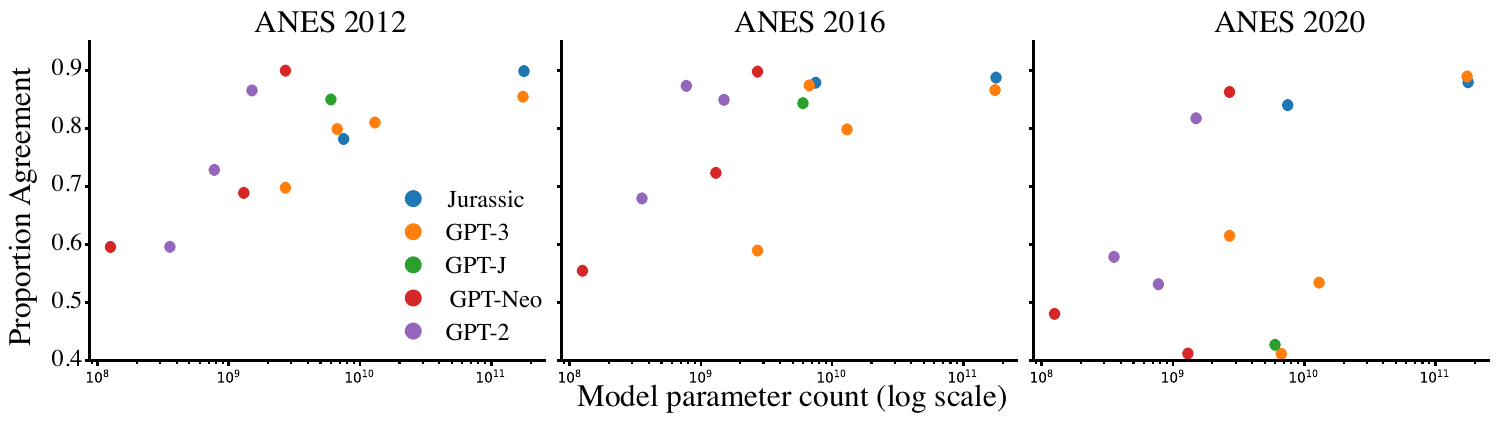}
\caption{Comparison of the performance of different language models on the vote prediction task. See text for details. }
\label{fig:model_comparison}
\end{figure*}

Finally, we tested the performance of different language models on the vote prediction task; results are shown in Figure~\ref{fig:model_comparison}.  Five different families of language models were tested, representing the best available models at the time of writing.  GPT-3 and Jurassic are commercial models available only via a paid API; all other models have been publicly released. The figure shows performance as a function of parameter count, but performance also depends strongly on the details of the corpus used to train each model.  We here only note that, as in so many other natural language processing tasks, the very largest models perform very well.  Also of note is the surprisingly good performance of the largest member of the GPT-Neo family of models - at only 6B parameters, its performance rivals that of the much larger and more costly GPT-3 (at 175B parameters).


\section{Details on Study 3}

\subsection{Data generation}
\begin{figure}
\begin{mdframed}
{\small{Interviewer: What is your gender? Are you ``male'' or ``female''?\\
Me: \underline{male}\\
Interviewer: I am going to read you a list of four race categories. What race do you consider yourself to be? ``White'', ``Black'', ``Asian'', or ``Hispanic''?\\
Me: \underline{white}\\
Interviewer: What is your age in years?\\
Me: \underline{29}\\
Interviewer: What is the highest level of school you have completed, or the highest degree you have received? Is it ``high school'', ``some college'', ``a four-year college degree'', or ``an advanced degree''?\\
Me: \underline{high school}\\
Interviewer:  When you see the American flag flying, how does it make you feel? Does it make you feel ``extremely good'', ``moderately good'', ``a little good'', ``neither good nor bad'', ``a little bad'', ``moderately bad'', or ``extremely bad''?\\
Me: \underline{extremely good}\\
Interviewer: Do you ever discuss politics with your family and friends? Please respond with ``yes'' or ``no''.\\
Me: \underline{yes}\\
Interviewer: How interested would you say you are in politics? Are you ``very interested'', ``somewhat interested'', ``not very interested'', or ``not at all interested''?\\
Me: \underline{somewhat interested}\\
Interviewer: Which would you say best describes your partisan identification. Would you say you are a ``strong democrat'', ``not very strong democrat'', ``independent, but closer to the Democratic party'', ``independent'', ``independent, but closer to the Republican party'', ``not very strong Republican'', or ``strong Republican''?\\
Me: \underline{strong Republican}\\
Interviewer: Did you vote in the 2016 general election? Please answer with ``yes'' or ``no''.\\
Me: \underline{yes}\\
Interviewer: Which presidential candidate did you vote for in the 2016 presidential election, ``Hillary Clinton'', ``Donald Trump'', or ``someone else''?\\
Me: \underline{Donald Trump}\\
Interviewer: Lots of things come up that keep people from attending religious services even if they want to. Thinking about your life these days, do you ever attend religious services? Please respond with ``yes'' or ``no''.\\
Me: \textbf{yes}
}}
\end{mdframed}
\caption{An interview-style context used in Study 3. The context is in plaintext; underline text shows demographic variables dynamically inserted into the interview template; one possible sampled completion is shown in bold.}
\label{fig:interview_example}
\end{figure}

For Study 3, we generated a silicon sample of virtual ANES participants by constructing a complete virtual interview. As the example template (Figure \ref{fig:interview_example}) indicates, we selected twelve common variables\footnote{Because vote choice is conditional on vote turnout, these two items are combined into one metric in the analysis, for a total of 11 unique items.} from the 2016 ANES for exploration, representing a variety of demographic (gender, race/ethnicity, age, education, religiosity), attitudinal (political interest, patriotic feelings about the flag, party identification, ideology), and behavioral (talk about politics, vote, and vote choice) information. The conditioning text included the mock interview with questions and responses for eleven of the twelve items, leaving the twelfth question for GPT-3 to answer. Each backstory was based on actual responses given by one human ANES respondent.\footnote{Missing responses in the ANES data resulted in the removal of the question from the conditioning text.} In study 2, the goal was to measure the probability of a single token; here the goal is to measure a wide variety of multi-token responses, which complicates the analysis of their raw probabilities.  Instead, we allow the GPT-3 API to sample completions, which we then analyze.

Like Study 2, we use a templating system that maps ANES demographic variables to template fragments, which are then concatenated to construct the conditioning context.  Because this was a virtual interview, we used phrasing that exactly matched the ANES interview verbiage whenever possible.  We mapped ANES variables to short text fragments, which were then interpolated into template fragments.
For this study, we used the following ANES variables (in the following order):
gender (V161342),
race (V161310x),
age (V161267),
education (V161270)
church attendance (V161244),
patriotism (V162125x),
whether the subject discusses politics (V162174),
level of interest in politics (V162256),
7-point self-reported ideology (V161126),
7-point self-reported partisanship (V161158x),
whether the subject voted in 2016 (V162031x) and for whom (V162062x). Table \ref{tab:questionwording} provides the full text of all ANES question wording and GPT-3 template text.

Recall that the goal of this study was to predict one factor (which we call the \emph{target factor}, such as ``Race'') given specific values of all of the other factors.  The template fragment for the target factor was always placed at the end of the context (and naturally did not include the corresponding ANES variable).  Given a context, we asked GPT-3 to sample 5 tokens, using a temperature of 0.7.  GPT-3's generated text was then lightly processed (lower-cased, stripped of leading and trailing whitespace), and compared to the limited set of ANES responses for the target factor using exact string matches.  So, for example, if the target question was ``Race'', then the string produced by GPT-3 would be compared to ``white'', ``black'', ``asian'' and ``hispanic'', and coded as 1, 2, 3 or 5, respectively.  If the GPT-3 response did not match any of the allowable responses, it was coded as missing data. This generates a dataset that is structurally equivalent to the original ANES dataset.

The combination of 12 variables and 4270 respondents resulted in the generation of more than 50,000 unique conditioning texts for GPT-3, each designed to elicit one ``silicon'' respondent's answer to one question. In our analysis, we keep only the 1782 observations that are complete in both ANES and GPT-3 responses. This prevents variation in the set of cases from introducing additional statistical noise to the comparison.

\newpage

\begin{longtable}{ccccc}
    \caption{Comparison of ANES question wording and GPT-3 Template} \\
    \hline
    Order & Variable & ANES VarID & ANES Question Wording & GPT-3 Template Text\\
    \hline
    \endfirsthead

    \multicolumn{5}{l}{\textit{...continued from last page.}}\\
    \hline
    Order & Variable & ANES VarID & ANES Question Wording & GPT-3 Template Text\\
    \hline
    \endhead

    \hline
    \multicolumn{5}{r}{\textit{continued on next page...}}\\
    \endfoot

    \hline
    \endlastfoot

    \multirow{3}{2em}{1} & \multirow{3}{4.5em}{Gender} & \multirow{3}{4em}{V161342} & \multirow{3}{12em}{What is your gender?} & \multirow{3}{12em}{What is your gender? Are you "male" or "female"?}\\&&&&\\&&&&\\
    \hline

    \multirow{14}{2em}{2} & \multirow{14}{4.5em}{Race / Ethnicity} & \multirow{14}{4em}{V161310x} & \multirow{14}{12em}{I am going to read you a list of five race categories. Please choose one or more races that you consider yourself to be: - white, - black or African- American, - American Indian or Alaska Native, - Asian, or - Native Hawaiian or other Pacific Islander? + Are you Spanish, Hispanic, or Latino?} & \multirow{14}{12em}{I am going to read you a list of four race categories. What race do you consider yourself to be? "White", "Black", "Asian", or "Hispanic"?}\\&&&&\\&&&&\\&&&&\\&&&&\\&&&&\\&&&&\\&&&&\\&&&&\\&&&&\\&&&&\\&&&&\\&&&&\\&&&&\\
    \hline

    \multirow{3}{2em}{3} & \multirow{3}{4.5em}{Age} & \multirow{3}{4em}{V161247} & \multirow{3}{12em}{(Derived variable - no question text)} & \multirow{3}{12em}{What is your age in years?}\\&&&&\\&&&&\\
    \hline

    \multirow{8}{2em}{4} & \multirow{8}{4.5em}{Education} & \multirow{8}{4em}{V161270} & \multirow{8}{12em}{What is the highest level of school you have completed or the highest degee you have received?} & \multirow{8}{12em}{What is the highest level of school you have completed, or the highest degree you have received? Is it "high school", "some college", "a four-year college degree", or "an advanced degree"?}\\&&&&\\&&&&\\&&&&\\&&&&\\&&&&\\&&&&\\&&&&\\
    \hline
    \pagebreak

    \multirow{11}{2em}{5} & \multirow{11}{4.5em}{Attends Church} & \multirow{11}{4em}{V161244} & \multirow{11}{12em}{Lots of things come up that keep people from attending religious services even if they want to. Thinking about your life these days, do you ever attend religious services, apart from occasional weddings, baptisms or funerals?} & \multirow{11}{12em}{Lots of things come up that keep people from attending religious services even if they want to. Thinking about your life these days, do you ever attend religious services? Please respond with "yes" or "no".}\\&&&&\\&&&&\\&&&&\\&&&&\\&&&&\\&&&&\\&&&&\\&&&&\\&&&&\\&&&&\\
    \hline

    \multirow{16}{2em}{6} & \multirow{16}{4.5em}{Patriotism} & \multirow{16}{4em}{V162125x} & \multirow{16}{12em}{When you see the American flag flying does it make you feel good, bad, or neither good nor bad? + Does it make you feel [extremely good, moderately good, or a little good / a little good, moderately good, or extremely good]? / Does it make you feel [extremely bad, moderately bad, or a little bad / a little bad, moderately bad, or extremely bad]?} & \multirow{16}{12em}{When you see the American flag flying, how does it make you feel? Does it make you feel "extremely good", "moderately good", "a little good", "neither good nor bad", "a little bad", "moderately bad", or "extremely bad"?}\\&&&&\\&&&&\\&&&&\\&&&&\\&&&&\\&&&&\\&&&&\\&&&&\\&&&&\\&&&&\\&&&&\\&&&&\\&&&&\\&&&&\\&&&&\\
    \hline

    \multirow{5}{2em}{7} & \multirow{5}{4.5em}{Discusses Politics} & \multirow{5}{4em}{V162174} & \multirow{5}{12em}{Do you ever discuss politics with your family or friends?} & \multirow{5}{12em}{Do you ever discuss politics with your family and friends? Please respond with "Yes" or "No".}\\&&&&\\&&&&\\&&&&\\&&&&\\
    \hline

    \pagebreak

    \multirow{11}{2em}{8} & \multirow{11}{4.5em}{Political Interest} & \multirow{11}{4em}{V162256} & \multirow{11}{12em}{How interested would you say you are in politics? Are you [very interested, somewhat interested, not very interested, or not at all interested / not at all interest, not very interested, somewhat interested, or very interested]?} & \multirow{11}{12em}{How interested would you say you are in politics? Are you "very interested", "somewhat interested", "not very interested", or "not at all interested"?}\\&&&&\\&&&&\\&&&&\\&&&&\\&&&&\\&&&&\\&&&&\\&&&&\\&&&&\\&&&&\\
    \hline

    \multirow{20}{2em}{9} & \multirow{20}{4.5em}{Voted in 2016} & \multirow{20}{4em}{V162031x} & \multirow{20}{12em}{In talking to people about elections, we often find that a lot of people were not able to vote because they weren’t registered, they were sick, or they just didn’t have time. Which of the following statements best describes you: One, I did not vote (in the election this November), Two, I thought about voting this time, but didn’t, Three, I usually vote, but didn’t this time, or Four, I am sure I voted? + (Derived from other Pre and Post Election Questions)} & \multirow{20}{12em}{Did you vote in the 2016 general election? Please answer with "yes" or "no".}\\&&&&\\&&&&\\&&&&\\&&&&\\&&&&\\&&&&\\&&&&\\&&&&\\&&&&\\&&&&\\&&&&\\&&&&\\&&&&\\&&&&\\&&&&\\&&&&\\&&&&\\&&&&\\&&&&\\
    \hline
    \pagebreak

    \multirow{9}{2em}{10} & \multirow{9}{4.5em}{2016 Vote Choice} & \multirow{9}{4em}{V162062x} & \multirow{9}{12em}{Who did you vote for? [Hillary Clinton, Donald Trump / Donald Trump, Hillary Clinton], Gary Johnson, Jill Stein, or someone else? + (Derived from other Pre and Post Election Questions)} & \multirow{9}{12em}{Which presidential candidate did you vote for in the 2016 presidential election, "Hillary Clinton", "Donald Trump", or "someone else"? \emph{Note: Only displayed if respondent voted.}}\\&&&&\\&&&&\\&&&&\\&&&&\\&&&&\\&&&&\\&&&&\\&&&&\\
    \hline

    \multirow{9}{2em}{11} & \multirow{9}{4.5em}{Ideology} & \multirow{9}{4em}{V161126} & \multirow{9}{12em}{Where would you place yourself on this scale, or haven’t you thought much about this? (Scale card shown or online response options)} & \multirow{9}{12em}{When asked about your political ideology, would you say you are "extremely liberal", "liberal", "slightly liberal", "moderate", "slightly conservative", "conservative", or "extremely conservative"?}\\&&&&\\&&&&\\&&&&\\&&&&\\&&&&\\&&&&\\&&&&\\&&&&\\
    \hline

    \multirow{15}{2em}{12} & \multirow{15}{4.5em}{Party ID} & \multirow{15}{4em}{V161158x} & \multirow{15}{12em}{Generally speaking, do you usually think of yourself as [a Democrat, a Republican / a Republican, a Democrat], an independent, or what? +  Would you call yourself a strong [Democrat / Republican] or a not very strong [Democrat / Republican]? OR Do you think of yourself as closer to the Republican Party or to the Democratic Party?} & \multirow{15}{12em}{Which would you say best describes your partisan identification. Would you say you are a "strong democrat", "not very strong democrat", "independent, but closer to the Democratic party", "independent", "independent, but closer to the Republican party", "not very strong Republican", or "strong Republican"?}\\&&&&\\&&&&\\&&&&\\&&&&\\&&&&\\&&&&\\&&&&\\&&&&\\&&&&\\&&&&\\&&&&\\&&&&\\&&&&\\&&&&
    \label{tab:questionwording}
\end{longtable}

\newpage
\subsection{Data analysis}
The complete set of synthetic responses are appended together to create a single dataset that includes the ANES values for all eleven variables and the silicon responses for all eleven variables.

As an important methodological note, we do not calculate the direct individual-level correspondence between the ANES value for a given respondent and the GPT-3 value based on the same backstory information (such as a percent correctly predicted). GPT-3 draws tokens from a distribution of words, and we also assume distributions in outcomes in the general population. Therefore, even if GPT-3 and ANES values are drawn from the same distribution, we cannot expect them to match in any given case. The important demonstration for our point is not whether GPT-3 can correctly predict an individual, but rather whether it can produce a distribution of generated responses that is comparable to the distribution in the human data.

We use the CramerV function of the R package `DescTools' to calculate the Cramer's V between every possible combination of the 22 variables. We use Cramer's V as it is amenable to calculation using categorical data, and, like Pearson's Chi-squared on which it is based, relies on marginal values to account for variations in base rates. Cramer's V has a range of 0 to 1. Higher values of Cramer's V indicate that knowing the value of one variable gives you more information about the likely value of the second variable.

Tables \ref{tab:cramersvnumbers1} - \ref{tab:cramersvnumbers3} report the Cramer's V values for Figure 6 in the main text of the paper.

\begin{table}[!htbp] \centering
\renewcommand{\arraystretch}{0.9}
\begin{tabular}{@{\extracolsep{5pt}} ccccc}
\toprule{}
ANES ``Input'' & ``Output'' Variable & ANES Cramer's V & GPT-3 Cramer's V & Difference \\
\hline \\[-1.8ex]
age & church.goer & 0.2 & 0.19 & 0.01 \\
age & discuss.politics & 0.21 & 0.21 & 0 \\
age & race & 0.21 & 0.2 & 0.01 \\
age & education & 0.25 & 0.21 & 0.04 \\
age & gender & 0.18 & 0.2 & -0.02 \\
age & ideology & 0.23 & 0.2 & 0.03 \\
age & patriotism & 0.21 & 0.21 & 0 \\
age & pid7 & 0.22 & 0.21 & 0.01 \\
age & political.interest & 0.22 & 0.21 & 0.01 \\
age & vote.2016 & 0.24 & 0.23 & 0.01 \\
church.goer & age & 0.2 & 0.2 & 0 \\
church.goer & discuss.politics & 0.01 & 0.14 & -0.13 \\
church.goer & race & 0.09 & 0.04 & 0.05 \\
church.goer & education & 0.06 & 0.01 & 0.05 \\
church.goer & gender & 0.04 & 0.02 & 0.02 \\
church.goer & ideology & 0.28 & 0.12 & 0.16 \\
church.goer & patriotism & 0.2 & 0.05 & 0.15 \\
church.goer & pid7 & 0.22 & 0.19 & 0.03 \\
church.goer & political.interest & 0.04 & 0.08 & -0.04 \\
church.goer & vote.2016 & 0.19 & 0.24 & -0.05 \\
discuss.politics & age & 0.21 & 0.22 & -0.01 \\
discuss.politics & church.goer & 0.01 & 0.18 & -0.17 \\
discuss.politics & race & 0.14 & 0.02 & 0.12 \\
discuss.politics & education & 0.2 & 0.11 & 0.09 \\
discuss.politics & gender & 0 & 0.08 & -0.08 \\
discuss.politics & ideology & 0.17 & 0.06 & 0.11 \\
discuss.politics & patriotism & 0.03 & 0.1 & -0.07 \\
discuss.politics & pid7 & 0.16 & 0.11 & 0.05 \\
discuss.politics & political.interest & 0.4 & 0.28 & 0.12 \\
discuss.politics & vote.2016 & 0.11 & 0.2 & -0.09 \\
race & age & 0.21 & 0.2 & 0.01 \\
race & church.goer & 0.09 & 0.07 & 0.02 \\
race & discuss.politics & 0.14 & 0.05 & 0.09 \\
race & education & 0.1 & 0.07 & 0.03 \\
race & gender & 0.08 & 0.07 & 0.01 \\
race & ideology & 0.12 & 0.1 & 0.02 \\
race & patriotism & 0.17 & 0.08 & 0.09 \\
race & pid7 & 0.18 & 0.1 & 0.08 \\
race & political.interest & 0.06 & 0.11 & -0.05 \\
race & vote.2016 & 0.17 & 0.11 & 0.06 \\
\bottomrule{}
\end{tabular}
  \caption{Cramer's V values}
  \label{tab:cramersvnumbers1}
\end{table}

\begin{table}[!htbp] \centering

\renewcommand{\arraystretch}{0.9}
\begin{tabular}{@{\extracolsep{5pt}} ccccc}
\toprule{}
ANES ``Input'' & ``Output'' Variable & ANES Cramer's V & GPT-3 Cramer's V & Difference \\
\hline \\[-1.8ex]

education & age & 0.25 & 0.23 & 0.02 \\
education & church.goer & 0.06 & 0.05 & 0.01 \\
education & discuss.politics & 0.2 & 0.07 & 0.13 \\
education & race & 0.1 & 0.04 & 0.06 \\
education & gender & 0.04 & 0.05 & -0.01 \\
education & ideology & 0.13 & 0.09 & 0.04 \\
education & patriotism & 0.09 & 0.05 & 0.04 \\
education & pid7 & 0.11 & 0.08 & 0.03 \\
education & political.interest & 0.12 & 0.07 & 0.05 \\
education & vote.2016 & 0.14 & 0.09 & 0.05 \\
gender & age & 0.18 & 0.21 & -0.03 \\
gender & church.goer & 0.04 & 0.01 & 0.03 \\
gender & discuss.politics & 0 & 0.01 & -0.01 \\
gender & race & 0.08 & 0.03 & 0.05 \\
gender & education & 0.04 & 0.07 & -0.03 \\
gender & ideology & 0.13 & 0.14 & -0.01 \\
gender & patriotism & 0.06 & 0.07 & -0.01 \\
gender & pid7 & 0.16 & 0.1 & 0.06 \\
gender & political.interest & 0.12 & 0.04 & 0.08 \\
gender & vote.2016 & 0.09 & 0.11 & -0.02 \\
ideology & age & 0.23 & 0.2 & 0.03 \\
ideology & church.goer & 0.28 & 0.07 & 0.21 \\
ideology & discuss.politics & 0.17 & 0.08 & 0.09 \\
ideology & race & 0.12 & 0.09 & 0.03 \\
ideology & education & 0.13 & 0.1 & 0.03 \\
ideology & gender & 0.13 & 0.12 & 0.01 \\
ideology & patriotism & 0.22 & 0.15 & 0.07 \\
ideology & pid7 & 0.37 & 0.32 & 0.05 \\
ideology & political.interest & 0.15 & 0.14 & 0.01 \\
ideology & vote.2016 & 0.4 & 0.28 & 0.12 \\
patriotism & age & 0.21 & 0.18 & 0.03 \\
patriotism & church.goer & 0.2 & 0.11 & 0.09 \\
patriotism & discuss.politics & 0.03 & 0.08 & -0.05 \\
patriotism & race & 0.17 & 0.1 & 0.07 \\
patriotism & education & 0.09 & 0.07 & 0.02 \\
patriotism & gender & 0.06 & 0.09 & -0.03 \\
patriotism & ideology & 0.22 & 0.14 & 0.08 \\
patriotism & pid7 & 0.19 & 0.16 & 0.03 \\
patriotism & political.interest & 0.08 & 0.17 & -0.09 \\
patriotism & vote.2016 & 0.25 & 0.15 & 0.1 \\

\bottomrule{}
\end{tabular}
  \caption{Cramer's V values}
  \label{tab:cramersvnumbers2}
\end{table}

\begin{table}[!htbp] \centering

\begin{tabular}{@{\extracolsep{5pt}} ccccc}
\toprule{}
ANES ``Input'' & ``Output'' Variable & ANES Cramer's V & GPT-3 Cramer's V & Difference \\
\hline \\[-1.8ex]
pid7 & age & 0.22 & 0.21 & 0.01 \\
pid7 & church.goer & 0.22 & 0.07 & 0.15 \\
pid7 & discuss.politics & 0.16 & 0.13 & 0.03 \\
pid7 & race & 0.18 & 0.07 & 0.11 \\
pid7 & education & 0.11 & 0.12 & -0.01 \\
pid7 & gender & 0.16 & 0.12 & 0.04 \\
pid7 & ideology & 0.37 & 0.32 & 0.05 \\
pid7 & patriotism & 0.19 & 0.15 & 0.04 \\
pid7 & political.interest & 0.12 & 0.16 & -0.04 \\
pid7 & vote.2016 & 0.48 & 0.37 & 0.11 \\
political.interest & age & 0.22 & 0.2 & 0.02 \\
political.interest & church.goer & 0.04 & 0.13 & -0.09 \\
political.interest & discuss.politics & 0.4 & 0.16 & 0.24 \\
political.interest & race & 0.06 & 0.04 & 0.02 \\
political.interest & education & 0.12 & 0.07 & 0.05 \\
political.interest & gender & 0.12 & 0.11 & 0.01 \\
political.interest & ideology & 0.15 & 0.1 & 0.05 \\
political.interest & patriotism & 0.08 & 0.16 & -0.08 \\
political.interest & pid7 & 0.12 & 0.12 & 0 \\
political.interest & vote.2016 & 0.12 & 0.12 & 0 \\
vote.2016 & age & 0.24 & 0.23 & 0.01 \\
vote.2016 & church.goer & 0.19 & 0.19 & 0 \\
vote.2016 & discuss.politics & 0.11 & 0.23 & -0.12 \\
vote.2016 & race & 0.17 & 0.07 & 0.1 \\
vote.2016 & education & 0.14 & 0.14 & 0 \\
vote.2016 & gender & 0.09 & 0.19 & -0.1 \\
vote.2016 & ideology & 0.4 & 0.34 & 0.06 \\
vote.2016 & patriotism & 0.25 & 0.16 & 0.09 \\
vote.2016 & pid7 & 0.48 & 0.37 & 0.11 \\
vote.2016 & political.interest & 0.12 & 0.2 & -0.08 \\
\bottomrule{}
\end{tabular}
  \caption{Cramer's V values}
  \label{tab:cramersvnumbers3}
\end{table}

\subsubsection{Missing Data}
In all presented analysis, the data are restricted to just the cases that are complete - meaning there are valid response values for all ANES \textit{and} GPT-3 variables. Of the 4270 cases in the 2016 ANES data file, 1782 are complete cases used in the analysis.

Table \ref{tab:naproportions} displays the percent of cases with missing data for each variable. The percent of missing data produced varies substantially, for both humans and GPT-3. GPT-3 was able to produce a valid and compliant answer in more than three-quarters of the cases for all question items, and three of the items had compliance rates above 99 percent. With additional training for particular questions, non-compliance could likely be reduced.

\begin{table}[!htbp] \centering

\begin{tabular}{@{\extracolsep{5pt}} ccc}
\toprule{}
 Variable & ANES & GPT-3 \\
\hline \\[-1.8ex]
 Age & 4.7 & 2.8 \\
 Attends Church & 0 & 0.4 \\
 Discusses Politics & 0.1 & 14.6 \\
 Race & 0.1 & 5.6 \\
 Education & 14.3 & 1 \\
 Gender & 0 & 1.2 \\
 Ideology & 4.2 & 22.6 \\
 Patriotism & 0.7 & 14.6 \\
 Party ID & 3.6 & 0.5 \\
 Political Interest & 1.1 & 14.8 \\
 2016 Vote and Choice & 0.5 & 23.8 \\
\bottomrule{}
\end{tabular}
  \caption{Percent of Observations Coded as Missing}
  \label{tab:naproportions}
\end{table}

\subsubsection{Descriptive Statistics}
Table \ref{tab:gpt3descriptives} presents the descriptive statistics for the variables used in Study 3, separately by data source (ANES humans or GPT-3 silicon sample).

\begin{table}[!htbp] \centering
\renewcommand{\arraystretch}{0.9}
\begin{tabular}{@{\extracolsep{5pt}}lcccccccc}
\\[-1.8ex]\hline \\[-1.8ex]
Variable & Source & \multicolumn{1}{c}{N} & \multicolumn{1}{c}{Mean} & \multicolumn{1}{c}{St. Dev.} & \multicolumn{1}{c}{Min} & \multicolumn{1}{c}{Pctl(25)} & \multicolumn{1}{c}{Pctl(75)} & \multicolumn{1}{c}{Max} \\
\hline \\[-1.8ex]
Age & ANES & 1,782 & 50.143 & 17.557 & 18 & 35 & 64 & 90 \\
Age & GPT-3 & 1,782 & 35.486 & 12.589 & 0 & 27 & 41 & 99 \\
Attends Church & ANES & 1,782 & 0.583 & 0.493 & 0 & 0 & 1 & 1 \\
Attends Church & GPT-3 & 1,782 & 0.604 & 0.489 & 0 & 0 & 1 & 1 \\
Talks Politics & ANES  & 1,782 & 0.864 & 0.343 & 0 & 1 & 1 & 1 \\
Talks Politics & GPT-3 & 1,782 & 0.896 & 0.305 & 0 & 1 & 1 & 1 \\
Ideology & ANES  & 1,782 & 4.095 & 1.618 & 1 & 3 & 5 & 7 \\
Ideology & GPT-3 & 1,782 & 4.017 & 1.737 & 1 & 3 & 5 & 7 \\
Patriotism & ANES & 1,782 & 1.963 & 1.299 & 1 & 1 & 2 & 7 \\
Patriotism & GPT-3 & 1,782 & 1.456 & 0.902 & 1 & 1 & 2 & 7 \\
Party ID & ANES  & 1,782 & 3.857 & 2.197 & 1 & 2 & 6 & 7 \\
Party ID & GPT-3 & 1,782 & 4.429 & 2.169 & 1 & 2 & 6 & 7 \\
Political Interest & ANES & 1,782 & 1.990 & 0.791 & 1 & 1 & 2 & 4 \\
Political Interest & GPT-3 & 1,782 & 1.694 & 0.852 & 1 & 1 & 2 & 4 \\
White & ANES  & 1,782 & 0.803 & 0.398 & 0 & 1 & 1 & 1 \\
White & GPT-3 & 1,782 & 0.974 & 0.159 & 0 & 1 & 1 & 1 \\
Hispanic & ANES  & 1,782 & 0.089 & 0.285 & 0 & 0 & 0 & 1 \\
Hispanic & GPT-3 & 1,782 & 0.001 & 0.033 & 0 & 0 & 0 & 1 \\
Asian & ANES & 1,782 & 0.030 & 0.171 & 0 & 0 & 0 & 1 \\
Asian & GPT-3 & 1,782 & 0.002 & 0.041 & 0 & 0 & 0 & 1 \\
Black & ANES & 1,782 & 0.077 & 0.267 & 0 & 0 & 0 & 1 \\
Black & GPT-3 & 1,782 & 0.023 & 0.150 & 0 & 0 & 0 & 1 \\
Some College & ANES & 1,782 & 0.348 & 0.476 & 0 & 0 & 1 & 1 \\
Some College & GPT-3 & 1,782 & 0.642 & 0.480 & 0 & 0 & 1 & 1 \\
Graduate Degree & ANES & 1,782 & 0.196 & 0.397 & 0 & 0 & 0 & 1 \\
Graduate Degree & GPT-3 & 1,782 & 0.002 & 0.041 & 0 & 0 & 0 & 1 \\
Bachelor's Degree & ANES  & 1,782 & 0.280 & 0.449 & 0 & 0 & 1 & 1 \\
Bachelor's Degree & GPT-3 & 1,782 & 0.269 & 0.443 & 0 & 0 & 1 & 1 \\
High School & ANES  & 1,782 & 0.176 & 0.381 & 0 & 0 & 0 & 1 \\
High School & GPT-3 & 1,782 & 0.088 & 0.283 & 0 & 0 & 0 & 1 \\
Male & ANES & 1,782 & 0.481 & 0.500 & 0 & 0 & 1 & 1 \\
Male & GPT-3 & 1,782 & 0.759 & 0.428 & 0 & 1 & 1 & 1 \\
Voted in 2016 & ANES & 1,782 & 0.871 & 0.335 & 0 & 1 & 1 & 1 \\
Voted in 2016 & GPT-3 & 1,782 & 0.832 & 0.374 & 0 & 1 & 1 & 1 \\
Trump Voter & ANES  & 1,553 & 0.438 & 0.496 & 0 & 0 & 1 & 1 \\
Trump Voter & GPT-3 & 1,483 & 0.245 & 0.430 & 0 & 0 & 0 & 1 \\
Clinton Voter & ANES  & 1,553 & 0.484 & 0.500 & 0 & 0 & 1 & 1 \\
Clinton Voter & GPT-3& 1,483 & 0.233 & 0.423 & 0 & 0 & 0 & 1 \\
Other Voter & ANES & 1,553 & 0.078 & 0.268 & 0 & 0 & 0 & 1 \\
Other Voter & GPT-3 & 1,483 & 0.523 & 0.500 & 0 & 0 & 1 & 1 \\
\bottomrule{}
\end{tabular}
  \caption{Study 3 Descriptive Statistics for ANES and GPT-3 Data}
  \label{tab:gpt3descriptives}
\end{table}

\subsection{Alternative Specifications}

\subsubsection{Completely Synthetic Data}

The data generation process results in one vector of synthetic data based on the ANES inputs for the other eleven items. When these synthetic vectors are combined, the result is a complete dataset of synthetic data. In the main text of the paper, the Cramer's V is calculated using the ANES ``input'' variable and the GPT-3 output. This provides the most direct comparison between the ANES and GPT-3 results, as they are both based on the same values for one half of the Cramer's V calculations.

\begin{figure}[t!]
\centering
\includegraphics[width=0.93\linewidth]{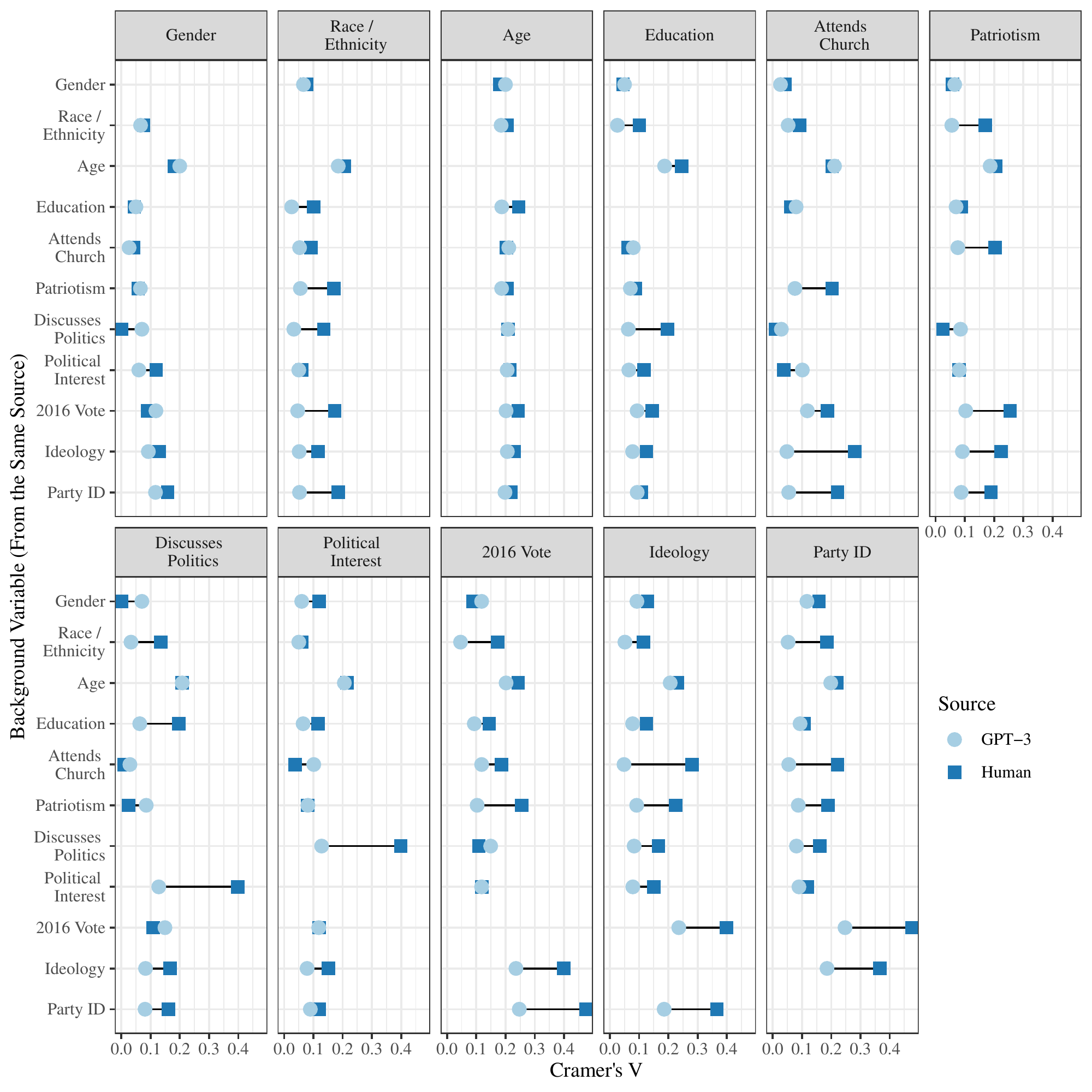}
\caption{Cramer's V Correlations in ANES vs. GPT-3 Data, using Entirely Synthetic GPT-3 Calculations}
\label{fig:gpt3ingpt3out}
\end{figure}

However, we can also estimate the Cramer's V between the various synthetic vectors, removing ANES data from the GPT-3 relationship calculation entirely. Figure \ref{fig:gpt3ingpt3out} shows the same data for the ``Human'' responses, but replaces the Cramer's V between ANES and GPT-3 that forms the ``GPT-3'' response in the main text with a Cramer's V calculation based entirely on synthetic data. Even though the use of synthetic data in both parts introduces additional noise in the estimation, the pattern of Cramer's V comparisons is highly similar to that seen when ANES inputs are used.

\subsubsection{GPT-3 Temperature Variation}

Additionally, when generating the GPT-3 results, the temperature setting can be varied. Temperature controls the amount of random variation allowed in the text sampling process used by GPT-3. In the main text, we use the industry standard temperature of .7. However, to demonstrate that the results are robust to multiple samples using different settings, we also provide a replication using temperature settings of 0.001 and 1.0. A setting of 0.001 means that in virtually all completions the algorithm will provide the response with the highest probability (meaning for a .49 to .51 split, all completions would return the token associated with the .51 probability). A setting of 1.0 means that the probability of selecting any particular token is equivalent to the probability distribution (i.e. there is no adjustment).

\begin{table}[!htbp] \centering
\begin{tabular}{@{\extracolsep{5pt}} cccc}
\toprule{}
 Summary Statistic & Temp: 0.001 & Temp: 0.7 & Temp: 1.0 \\
\hline \\[-1.8ex]
 Mean & 0.059 & -0.026 & -0.031 \\
 Minimum & -0.123 & -0.241 & -0.250 \\
 Maximum & 0.700 & 0.168 & 0.119 \\
 Standard Deviation & 0.141 & 0.068 & 0.070 \\
 N & 2518 & 1782 & 1022 \\
\bottomrule{}
\end{tabular}
  \caption{Average Error in Cramer's V Based on Varying Temperatures}
  \label{tab:temp_comparisons}
\end{table}


Table \ref{tab:temp_comparisons} shows summary statistics of the difference in Cramer's V between human and GPT-3 produced responses based on varying temperature settings.  In other words, the value for Cramer's V produced with human data is subtracted from the value for Cramer's V produced with GPT-3 data. These results mirror the main text, and are based on ANES "inputs" and GPT-3 "outputs." Summary statistics are then calculated based on the differences. We see that, of the three options, a temperature setting of .7 produces the lowest difference between Human and GPT-3 relationships. There are minimal differences between a temperature of .7 and a temperature of 1.0. We ran each temperature query once, and did not select the presented models for best fit from a range of probabilistic runs. The results provide some evidence that the relationship patterns uncovered by GPT-3 are robust to variations in model specification.

A temperature of 0.001 produces more error and also systematically overstates the relationship between the human input and GPT-3 output variables. One caveat: at a temperature of 0.001, GPT-3 identified all respondents as white. Without variation in this variable, we were unable to calcualte Cramer's V, so race / ethnicity is excluded from the calculations for a temperature of 0.001.

All Cramer's V calculations use the set of cases that have non-missing data for all human and GPT-3 produced variables. The different temperature settings produce a different number of valid completed cases. Lower temperature is more deterministic, and so minimizes the number of invalid tokens used as text completions. Higher temperatures sample from a range of tokens that includes more invalid responses. Therefore, mid-range temperatures appear to produce the desirable balance between validity and completeness.


\section{Cost Analysis}

The GPT-3 and Jurassic models are available only through a paid API. In the interests of full transparency, we here report the costs for Studies 1, 2 and 3. We only report costs for the final runs, but note that additional runs were performed as part of the experimental rhythm.

Study 1 required 1,1471 model queries (one for each human subject). The backstories were relatively small, at an average of 66 tokens. For each query, we generated a maximum of 128 tokens from the model. At the standard rate of \$0.06 / 1,000 tokens, this experiment cost a total of \$29.

Study 2 consisted of 3 experiments, one for ANES 2012, 2016 and 2020. We ran one model query per participant (5,914 in 20112, 4,270 in 2016 and 5,442 in 2020), for a total of 15,626 queries. Backstories were a bit longer than in Study 1, at an average of 80 tokens, but we only needed to generate one token per query, incurring a total cost of \$75.

Study 3 was more expensive. Because of the extended interview format, each prompt required an average of 458 tokens. For each query, we generated a maximum of 5 tokens.  We performed one query for each ANES participant, for a total of 4,270 queries, resulting in a total per experiment of \$119.  However, recall that Study 3 involved 12 different experiments (systematically predicting one backstory element from the others), and so the total cost of Study 3 was \$1,428.

We briefly note that using the Jurassic model (not available at the time of writing) would have reduced costs for all experiments.

\section*{Acknowledgements}
The authors thank the departments of Computer Science and Political Science at Brigham Young University for support and for feedback on this manuscript. Additionally, Chris Bail, Sarah Shugars, Luwei Ying, Diogo Ferrari, and Christopher Karpowitz provided invaluable feedback.




\bibliographystyle{unsrtnat}
\bibliography{refs}

\end{document}